\documentclass[sigconf]{acmart}
\usepackage{amsthm,multirow}
\usepackage{mathtools}
\usepackage{booktabs}
\usepackage{amsmath}
\usepackage{array,enumitem}
\usepackage[linesnumbered,ruled]{algorithm2e}
\usepackage{bm,changepage}
\usepackage{subcaption,siunitx,booktabs}
\newcommand{\camel}{\textsf{Camel}}
\newcommand{\proposed}{\textsf{TaPEm}}
\newcommand*{\affmark}[1][*]{\textsuperscript{#1}}
\def\BibTeX{{\rm B\kern-.05em{\sc i\kern-.025em b}\kern-.08emT\kern-.1667em\lower.7ex\hbox{E}\kern-.125emX}}
\setcopyright{acmcopyright}
\begin{document}

%
% The "title" command has an optional parameter, allowing the author to define a "short title" to be used in page headers.
%\title{Deep Contextualized Pair Representations for Task-Guided Heterogeneous Network Embedding}
\title{Task-Guided Pair Embedding in Heterogeneous Network}

\author{Chanyoung Park\affmark[1], Donghyun Kim\affmark[2], Qi Zhu\affmark[1], Jiawei Han\affmark[1], Hwanjo Yu\affmark[3]*}\thanks{*Corresponding Author}
\affiliation{%
	\institution{\affmark[1]University of Illinois at Urbana-Champaign, IL, USA, \affmark[2]Yahoo! Research, CA, USA}
}
%\affiliation{%
%	\institution{\affmark[2]Yahoo! Research, CA, USA}
%}
\affiliation{%
	\institution{\affmark[3]Pohang University of Science and Technology (POSTECH), Pohang, South Korea}
}
\email{{pcy1302, qiz3, hanj}@illinois.edu, donghyun.kim@verizonmedia.com, hwanjoyu@postech.ac.kr}
%
% By default, the full list of authors will be used in the page headers. Often, this list is too long, and will overlap
% other information printed in the page headers. This command allows the author to define a more concise list
% of authors' names for this purpose.

%
% The abstract is a short summary of the work to be presented in the article.
\begin{abstract}
Many real-world tasks solved by heterogeneous network embedding methods can be cast as modeling the likelihood of a pairwise relationship between two nodes. 
For example, the goal of author identification task is to model the likelihood of a paper being written by an author (paper--author pairwise relationship).
Existing task-guided embedding methods are node-centric in that they simply measure the similarity between the node embeddings to compute the likelihood of a pairwise relationship between two nodes.
However, we claim that for task-guided embeddings, it is crucial to focus on \textit{directly} modeling the pairwise relationship.
%which is equivalent to modeling the paper--author pairwise relationship.
%For example, for author identification task, the goal is to model the likelihood of a paper being written by an author, which is equivalent to modeling the paper--author pairwise relationship.
In this paper, we propose a novel task-guided pair embedding framework in heterogeneous network, called~\proposed, that directly models the relationship between a pair of nodes that are related to a specific task (e.g., paper-author relationship in author identification). 
%Existing methods learn node embeddings that are guided to perform well on specific tasks, and compute the likelihood of the pairwise relationship by similarity between their node embeddings.
%However, 
%Unlike existing methods that focus on learning node embeddings that are guided to perform well on specific tasks, 
%we claim that for task-guided embedding, it is crucial to explicitly model the relationship between two target nodes of interest.
%To compute the likelihood of a pairwise relationship between two nodes, existing task-guided embedding methods simply measure the similarity between their embeddings.
%that are specifically guided to perform well on specific tasks. 
%However, we claim that for task-guided embeddings, it is crucial to focus on directly modeling the relationship between two nodes.
% of interest.
%, as the ultimate goal is to infer the likelihood of the pairwise relationship between two nodes.
%because many tasks can be cast as predicting the likelihood of the pairwise relation. 
To this end, we 1) propose to learn a pair embedding under the guidance of its associated context path, i.e., a sequence of nodes between the pair, and 2) devise the pair validity classifier to distinguish whether the pair is valid with respect to the specific task at hand.
By introducing pair embeddings that capture the semantics behind the pairwise relationships, we are able to learn the fine-grained pairwise relationship between two nodes, which is paramount for task-guided embedding methods.
Extensive experiments on author identification task demonstrate that~\proposed~outperforms the state-of-the-art methods, especially for authors with few publication records.
\end{abstract}

%\ccsdesc[500]{Computer systems organization~Embedded systems}

%
% Keywords. The author(s) should pick words that accurately describe the work being
% presented. Separate the keywords with commas.

\copyrightyear{2019} 
\acmYear{2019} 
\acmConference[CIKM '19]{The 28th ACM International Conference on Information and Knowledge Management}{November 3--7, 2019}{Beijing, China}
%\acmBooktitle{The 28th ACM International Conference on Information and Knowledge Management (CIKM '19), November 3--7, 2019, Beijing, China}
\acmPrice{15.00}
\acmDOI{10.1145/3357384.3357982}
\acmISBN{978-1-4503-6976-3/19/11}
%
% This command processes the author and affiliation and title information and builds
% the first part of the formatted document.
\maketitle
\section{Introduction}
The goal of network embedding is to learn low dimensional representations for nodes in a network while preserving the network structure~\cite{grover2016node2vec,perozzi2014deepwalk} and properties~\cite{huang2017label,goyal2018capturing,ou2016asymmetric,zhang2018link,wang2017community}. 
As a large number of social and information networks are heterogeneous in nature, i.e., nodes and edges are of multiple types, heterogeneous network embedding methods have recently garnered attention~\cite{sun2009ranking,sun2011pathsim,zhang2018metagraph2vec,sun2011co}.
They learn node embeddings by exploiting various types of relationships among nodes and the network structure, and use them for general downstream tasks, such as node classification~\cite{shi2018aspem,dong2017metapath2vec}, link prediction~\cite{shi2018easing,fu2017hin2vec}, and clustering~\cite{chang2015heterogeneous,hussein2018meta}.

Recent studies have shown that previous heterogeneous network embedding methods fall short of solving a specific task, such as anomaly detection~\cite{chen2016entity},  recommendation~\cite{shi2019heterogeneous}, sentiment link prediction~\cite{wang2018shine}, and author identification~\cite{chen2017task,zhang2018camel},  to name a few, because they learn general purpose node embeddings.
In this regard, recently proposed task-guided network embedding methods guide node embeddings to perform well on a specific task by introducing a task-specific objective, instead of learning general purpose node embeddings that preserve the overall proximity among nodes.
For example,  Shi et al., generate node sequences that are meaningful for recommendation by meta-path based random walk~\cite{dong2017metapath2vec}, and integrate them with matrix factorization to specifically optimize for the rating prediction task~\cite{shi2019heterogeneous}.
%Moreover, Chen et al., model pairwise interactions of heterogeneous entity types to directly calculate the likelihood of events being anomalous~\cite{chen2016entity}.
Moreover, Wang et al., integrate users' sentiment relation network, social relation network and profile knowledge network into a heterogeneous network to specifically optimize for the sentiment link prediction task~\cite{wang2018shine}.
%rather than only focusing on a single particular type of network.
%These methods guided the node embeddings to perform well on their specific tasks by introducing task-guided loss, instead of learning general purpose node embeddings that preserve overall proximity and property among nodes, 

However, for task-guided embedding methods, it is crucial to particularly focus on \textit{directly modeling the pairwise relationship between two nodes}, because their ultimate goal is usually to model the likelihood of the pairwise relationship. i.e., the link probability between two nodes.
%i.e., explicit modeling of the link between the two target nodes on which we aim to infer the relationship.  
For example, for recommendation, the goal is to model the likelihood of a user favoring an item (i.e., user--item pairwise relationship). For author identification, the goal is to model the likelihood of a paper being written by an author (i.e., paper--author pairwise relationship). 
Nevertheless, previous task-guided embedding methods are node-centric in that they learn task-guided \textit{node embeddings}, and then simply compute the likelihood of a pairwise relationship between two nodes by employing a similarity metric, such as inner product~\cite{shi2019heterogeneous,wang2018shine,chen2016entity} or Euclidean distance~\cite{zhang2018camel}, between the learned node embeddings.
%, which we refer to as node-centric approach.

%are based on the node-centric assumption in which task-guided \textit{node embeddings} are first learned, followed by computing the likelihood between two nodes by measuring a similarity measure, such as inner product~\cite{shi2019heterogeneous,wang2018shine,chen2016entity} or Euclidean distance~\cite{zhang2018camel}, between their learned node embeddings.
%Nevertheless, previous task-guided embedding methods are based on the node-centric assumption in which task-guided \textit{node embeddings} are first learned, followed by computing the likelihood between two nodes by measuring a similarity measure, such as inner product~\cite{shi2019heterogeneous,wang2018shine,chen2016entity} or Euclidean distance~\cite{zhang2018camel}, between their learned node embeddings.

%whereas the ultimate end task is to model the relation (i.e., edge) between two nodes. In the end, these methods compute the likelihood between two nodes by measuring a similarity measure, such as inner product~\cite{shi2019heterogeneous,wang2018shine,chen2016entity} or Euclidean distance~\cite{zhang2018camel}, between their learned node embeddings.

In the light of this issue,
%the issue regarding the node-centric assumption, 
we propose a novel \textsf{\textbf{Ta}}sk-guided \textsf{\textbf{P}}air \textsf{\textbf{Em}}bedding framework in heterogeneous network, called~\proposed, that directly embeds a pair of nodes whose likelihood we wish to model. i.e., task-guided \textit{pair embedding}. 
Unlike previous node-centric task-guided embedding methods, the task-specific objective of~\proposed~is derived from the pair embedding.
Our intuition is that if a pair embedding can capture the semantics behind the relationship between two nodes that constitute the pair, we can model the fine-grained pairwise relationship better than the case when each node only has a single embedding. 
%\textcolor{red}{We note that learning pair embeddings can be viewed as splitting each node embedding into multiple embeddings considering its relationship with multiple different nodes, which is in contrast to having a single embedding for each node.}
%We note that learning a pair embedding can be viewed as having multiple embeddings for each node, which is useful to model fine-grained pairwise relationship between two nodes compared with having a single embedding for each node. 
As an illustration, consider the following toy example.
%We note that a pair embedding is composed of two node embeddings, and thus it can be viewed as having multiple embeddings for a node, where each embedding considers the fine-grained pairwise relationship between the two nodes. As an illustrative example, we provide a Toy Example.

\begin{figure}[t]
	\centering
	\includegraphics[width=0.94\linewidth]{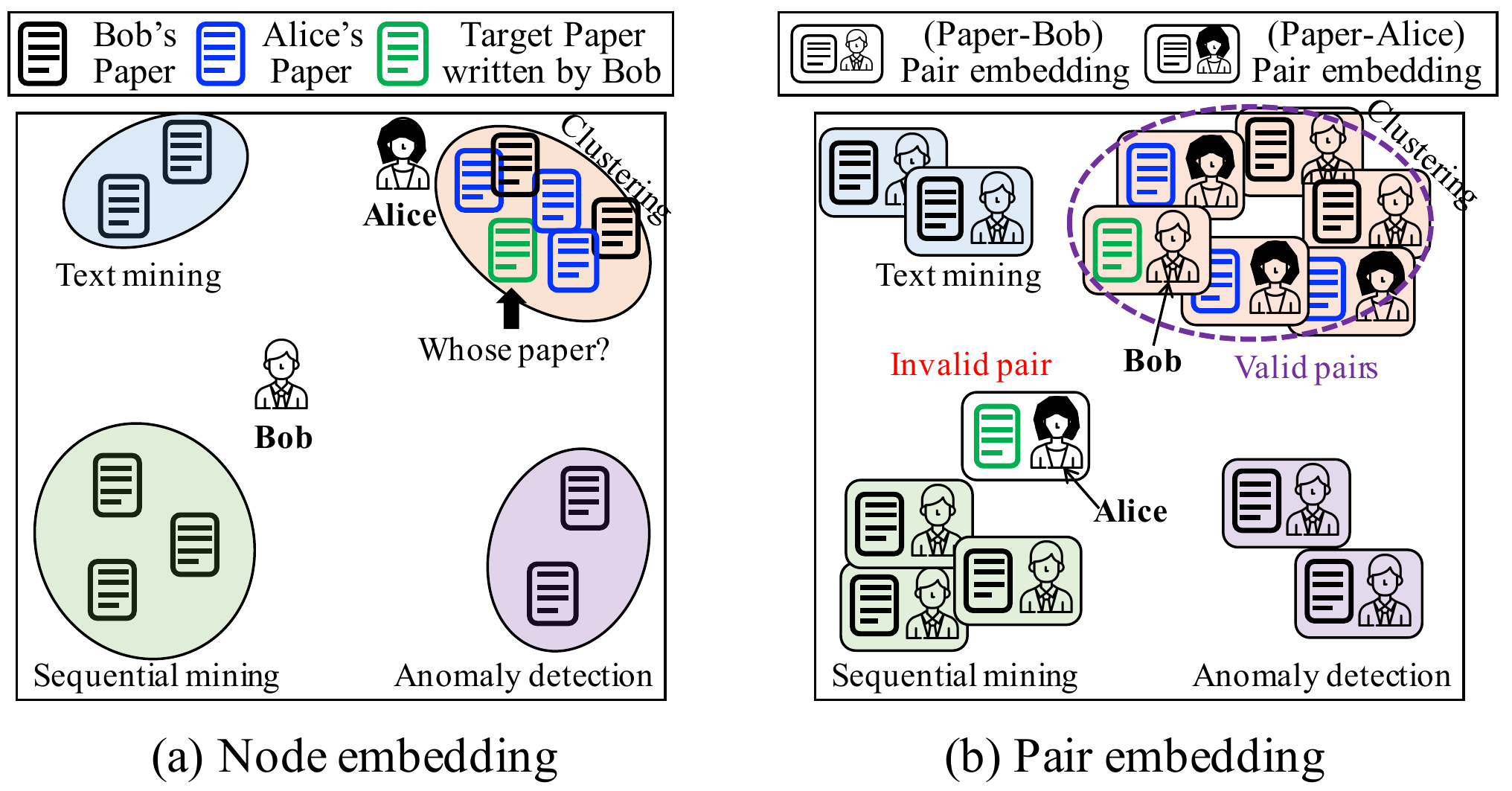}
	\vspace{-2ex}
	\caption{Comparisons of example visualization between node embedding and pair embedding.}
	\label{fig:author}
	\vspace{-2ex}
\end{figure}

\begin{adjustwidth}{5pt}{5pt}
\smallskip
\noindent\textbf{\textbf{\underline{Toy Example. }}}
Take an example of author identification scenario, which aims to find the authors of anonymous papers~\cite{zhang2018camel,chen2017task}. Figure~\ref{fig:author}(a) illustrates an example visualization of embeddings of papers and authors, where each author is assigned a single embedding.
We assume that Bob has written multiple papers in various research areas, whereas Alice's work are solely devoted to ``clustering''; hence, Alice is embedded close to papers whose topics are ``clustering''. Since each author has a single embedding, it has to be embedded to a single point that is optimal considering all of his diverse research areas. 
%it will try to find a globally optimal embedding that takes into account all of his diverse research areas. 
%In~\camel, papers with similar research topics are likely to be gathered together because the paper content is encoded. Thus,~\camel~would try to find a globally optimal embedding for Bob that takes into account all of his diverse research areas. 
However, a problem arises if we were to identify a true author of ``Target Paper'' about ``clustering'' (in green), which is written by Bob.
In this case, since Alice, who has written papers only on ``clustering'', is likely to be embedded closer to papers on ``clustering'' than Bob, Bob will be eventually ranked lower than Alice, which is not a desired result.
%~\footnote{Note that if Bob is an exceptionally productive author who has many publications on all of his diverse research areas, Bob may still be ranked higher than Alice. \textcolor{red}{But we are considering a general case in this toy example.}}.
%The only case in which Bob will be likely to be predicted as a true author of ``Target Paper'' is when Bob is a very active author who writes many papers in many different domains.
%\textcolor{red}{\textbf{in perspective of Bob}}
%\textcolor{red}{Alice who has written papers only on ``clustering'' will be ranked higher than Bob as a true author of ``Target paper'', because Alice is embedded closer to papers on ``clustering'', which is not a desired result.}
%In this case, Alice will be ranked higher than Bob as Alice has written papers only on ``clustering'', and thus embedded closer to papers on ``clustering'', which is not a desired result. 
%On the other hand, as shown in Figure~\ref{fig:author}(b), if we can embed each paper--author pair such that each pair embedding independently captures its associated research topic, (``Target paper'', Bob) pair can be embedded closer to the pairs related to ``clustering'' than invalid (``Target paper'', Alice) pair is.
On the other hand, as shown in Figure~\ref{fig:author}(b), if we can embed each paper--author pair such that each pair embedding independently captures its associated research topic and its validity information, (``Target paper'', Bob) pair can be embedded closer to the valid pairs related to ``clustering'' than invalid (``Target paper'', Alice) pair is.
%is embedded far from valid pairs as ``Target paper'' is not written by Alice
%\footnote{Hereafter, we will continue our discussions on the author identification scenario for the ease of explanation.}.
%Moreover, we want (``Target paper'', Alice) pair to be embedded far from valid pairs as ``Target paper'' is not written by Alice\footnote{Hereafter, we will continue our discussions on the author identification scenario for the ease of explanation.}.
%if we can embed each paper--author pair such that the pair embedding captures not only its associated research topic but also its validity, i.e., whether or not the paper is written by the author, Alice will not be blindly embedded close to pairs related to ``clustering'', and thus the ``Target paper'' will be correctly classified as a paper written by Bob.
%which is a desired outcome.
\end{adjustwidth}
%In short, we claim that authors should be embedded not only by globally considering their relevant research areas, but also more fine-grained local treatment is required.

In this regard, the key for successfully learning a task-guided pair embedding boils down to modeling 1) the semantics (e.g., research topic) behind the pairwise relationship, and 2) the validity of the pair regarding a specific task (e.g., given a paper--author pair, whether the paper in the pair is written by the author in the pair). Note that we will continue our discussions on the author identification scenario hereafter for the ease of explanation.

To capture the semantics behind the pairwise relationship, we propose to explicitly encode the paths between a paper--author pair,
%and try to predict the paths by using the pair embedding, assuming that the path reflects the research topic associated with it
%obtained from meta-path guided random walks on the academic heterogeneous network, 
where we denote such paths as \textit{context paths}.
%The core idea is to \textit{embed paper--author ``pairs'' that appear within the same context window of  random walks}.
%Figure~\ref{fig:pair} illustrates how paper--author pairs, and their associated context path are defined for a sample meta-path guided random walk shown in Figure~\ref{fig:author}(b). 
More precisely, from meta-path guided random walks~\cite{dong2017metapath2vec},
we first extract paper--author pairs each of which is associated with multiple context paths  (\textbf{Sec.~\ref{sec:het}}). Then, we embed each pair (\textbf{Sec.~\ref{sec:pair}}) and its associated context path (\textbf{Sec.~\ref{sec:path}}) into a vector, respectively. 
%Then,
%we find a representation for each pair that are useful for predicting its associated context path, 
Under the assumption that a context path reveals the research topic associated with the pair, we make the pair embedding naturally get similar to the embeddings of more frequently appearing context paths (\textbf{Sec.~\ref{sec:joint}}).
%predict the context path embedding vector given the pair embedding vector, 
%After obtaining the pairs and their context paths as above,~\proposed~predicts the context paths given the pair embedding, assuming that the context path reflects the research topic associated with the pair.
%In this way, the pair embedding will naturally get similar to the embeddings of more frequently appearing context paths, which characterize the research topic related to the pair. 
By encoding the related research topic into a paper--author pair embedding, we can model the fine-grained pairwise relationship between the paper--author pair.
%By doing so, a paper--author pair embedding can reflect the fine-grained pairwise relationship between the paper--author pair.
%We note that embedding paper--author pairs facilitates~\proposed~to model the fine-grained pairwise relationship between paper--author pair.
%More precisely, we encode each context path by using bidirectional gated recurrent unit (GRU) followed by attention module to extract a feature representation vector from each context path.
%Then, after obtaining the embeddings for paper--author pairs,
In the meantime, for each paper--author pair, we introduce a pair validity classifier to distinguish whether the pair is valid or not with respect to the specific task at hand (\textbf{Sec.~\ref{sec:validity}}).
%, i.e., whether the paper is written by the author.
%where a valid paper--author pair implies that the paper is written by the author. 
%Unlike skip-gram model based~\camel~that blindly makes the embeddings of papers and authors similar if they appear in the same context window whether or not the pairs are valid,  our pair validity classifier 
By reflecting the pair validity information into the pair embedding, we make two nodes that constitute a pair close to each other not only if they are related to a similar research topic, but also the pair itself is valid at the same time. The pair embeddings obtained from the above process eventually makes it easier to distinguish the valid pairs from the invalid ones.
%\textcolor{red}{This process encodes pair validity information into the pair embedding, which makes two pair embeddings that share a similar research topic to be close only if they are valid. }
%This is different from the skip-gram model based~\camel~that blindly makes the embeddings of papers and authors similar if they appear in the same context window, whether or not the pairs are valid.
%Eventually,~\proposed~is able to not only focuses on the accuracy of identifying well-known authors but also that of lesser known authors.

To verify the benefit of our task-guided pair embedding framework in heterogeneous network, we specifically focus on the task of \textbf{author identification} in big scholarly data~\cite{priem2013scholarship}. Our extensive experiments demonstrate that~\proposed~considerably outperforms the state-of-the-art task-guided heterogeneous embedding methods, especially for authors with few publication records (\textbf{Sec.~\ref{sec:exp}}).
We also perform various experiments to qualitatively ascertain the benefit of the pair embedding framework of~\proposed.

\section{Related Work}
%\subsection{Network Embedding}
\noindent\textbf{Network Embedding. }
The goal of network embedding is to learn a low dimensional representation for each node of a network while
preserving the network structure and various properties such as attributes related to nodes~\cite{huang2017label,huang2017accelerated,chang2015heterogeneous} and edges~\cite{goyal2018capturing}. The learned embeddings are then used for general downstream tasks, such as node classification~\cite{bhagat2011node}, link prediction~\cite{zhang2018link,ou2016asymmetric}, 
%recommendation~\cite{zhou2017scalable,wang2018graphgan}, 
and clustering~\cite{wang2017community}.
Inspired by the recent advancement of word embedding techniques in natural language processing~\cite{mikolov2013distributed}, numerous network embedding approaches based on random walk have been proposed~\cite{grover2016node2vec,perozzi2014deepwalk}. DeepWalk~\cite{perozzi2014deepwalk} and node2vec~\cite{grover2016node2vec} combine random walk and skip-gram to learn node embeddings. However, as these methods are proposed for homogeneous networks in which nodes and edges are of a single type, and thus are not suitable for modeling heterogeneous networks, there has been a line of research on heterogeneous network embedding~\cite{dong2017metapath2vec,wang2018shine,fu2017hin2vec,shi2018easing,zhang2018metagraph2vec}. Specifically, metapath2vec~\cite{dong2017metapath2vec} proposed a random walk scheme that is conditioned on meta-paths, and learned node embeddings by heterogeneous skip-gram with negative sampling. 
JUST~\cite{hussein2018meta} tackled the limitation of meta-path based random walk, and proposed random walks with jump and stay strategies.
Hin2Vec~\cite{fu2017hin2vec} considered the relationship between nodes to learn node embeddings, but it considered all possible relationships between two nodes aiming at learning general node embeddings, which is distinguished from our proposed framework in that we specifically focus on a single relationship related to a specific task.
Although these methods have been shown to be effective on general downstream tasks such as node classification, clustering and similarity search, they are not specifically designed to perform well on specific tasks such as recommendation~\cite{shi2019heterogeneous}, anomaly detection~\cite{chen2016entity}, sentiment link prediction~\cite{wang2018shine} and author identification~\cite{chen2017task,zhang2018camel}. 
We propose a novel task-guided pair embedding framework in heterogeneous network, and focus on the problem of author identification as an application of our framework. 
%We will explain the task in detail in the following subsection.

\begin{figure}[t]
	\centering
	\includegraphics[width=0.7\linewidth]{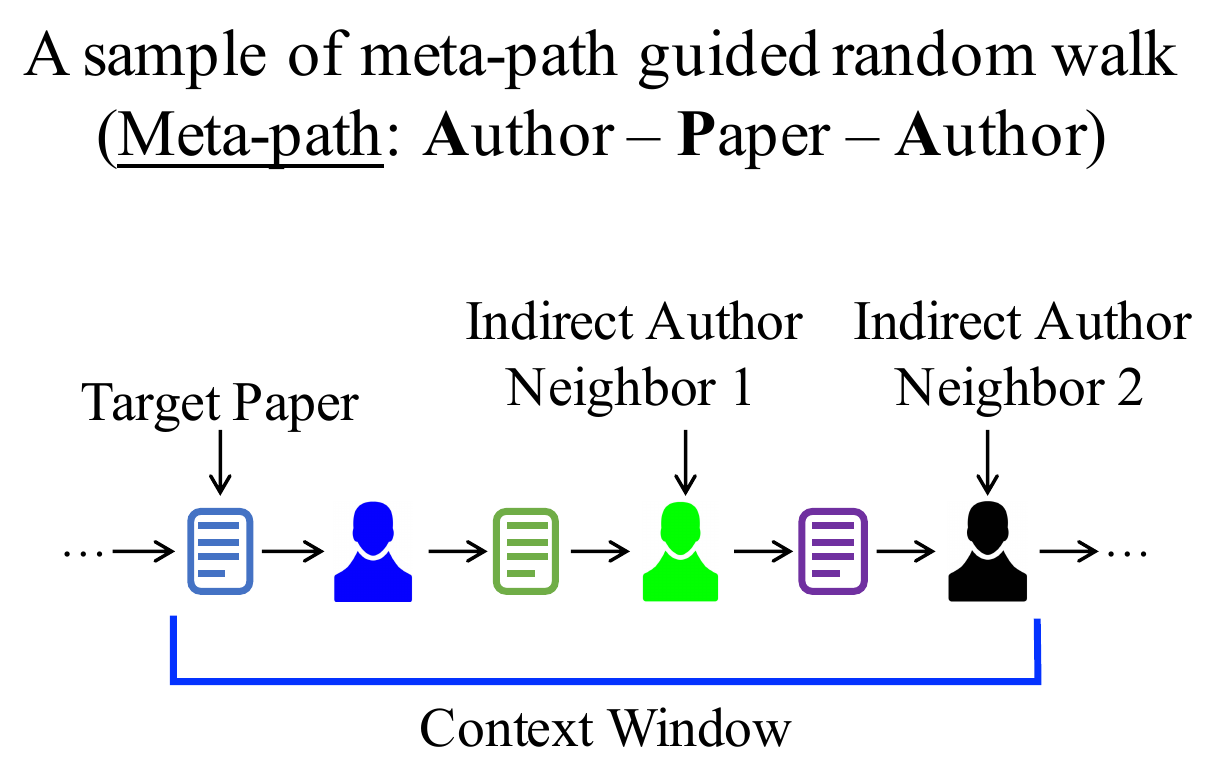}
%		\vspace{-2ex}
	\caption{Example showing the behavior of skip-gram model of~\camel~on a sample meta-path guided random walk.}
	\label{fig:camel_walk}
		\vspace{-1ex}
\end{figure}
\medskip
%\subsection{Author Identification}
\noindent\textbf{Author Identification. }
Many conferences in computer science adopt the double-blind review policy to eliminate bias in favor of well-known authors.
% Tomkins et al. showed that single-blind reviewers are significantly more likely than double-blind reviewers to recommend for acceptance papers from famous authors and top institutions~\cite{tomkins2017reviewer}. 
However, a knowledgeable reviewer can often disclose the authors of a paper by its content~\cite{chen2017task,zhang2018camel}. In this regard, the effectiveness of the double-blind review process is constantly being questioned by the research community~\cite{goues2017effectiveness,blank1991effects}. 

To specifically focus on the author identification task, Chen and Sun proposed a task-guided heterogeneous network embedding method, called HNE, that learns embeddings for authors, and different types of information of a paper, such as keywords, references and venues~\cite{chen2017task}. 
More recently, Zhang et al., proposed~\camel~\cite{zhang2018camel}~that encodes the semantic content of papers (i.e., abstract) instead of keywords, and considers indirectly correlated paper--author pairs obtained from meta-path guided random walk on the academic heterogeneous network using heterogeneous skip-gram model~\cite{dong2017metapath2vec}. Although~\camel~has shown its effectiveness, it suffers from an inherent limitation that each author has a single embedding even though authors may have diverse research areas (Refer to \textbf{Toy Example} in Figure~\ref{fig:author}(a)). Another limitation of~\camel~is that it 
%favors 
inadvertently makes a paper embedding and an author embedding similar to each other
%paper--author pairs that 
if they frequently appear together within a context window of a random walk sequence, whether or not the author is a true author of the paper. Figure~\ref{fig:camel_walk} shows a sample segment of a meta-path guided random walk. The skip-gram model of~\camel~trains the embedding of ``Target Paper'' to be similar to the embeddings of both Indirect Author Neighbor 1 and 2, regardless of their true authorship. i.e., whether or not they are the true authors of ``Target Paper''.
After performing multiple meta-path guided random walks, it is natural that an active author (i.e., an author with many publications) is more likely to appear frequently together with ``Target paper'' within the same context window than an inactive author. Therefore, according to the above skip-gram model, if an active author is a true author of ``Target paper'', then~\camel~can provide correct predictions. However, if an inactive author is a true author of ``Target paper'',~\camel~performs poorly because the inactive author does not appear frequently together with ``Target paper'', and thus not trained enough to be similar to ``Target paper''. To make the matter worse, if an active author is not a true author but appears frequently with ``Target paper'',~\camel~will still predict the active author as the true author.
% because he/she appears frequently with ``Target paper'' within the same context window.
In other words,~\textit{\camel~is biased to active authors}, which is a consequence of the skip-gram model.
%if an active author (i.e., an author with many publications) is a true author of ``Target paper'', it is more likely that the author appears frequently with ``Target paper'' in the same context window, compared with the case where an inactive author is a true author.
%In consequence,~\camel~expectedly shows high identification accuracy for active authors but performs rather poorly for relatively inactive authors, and the performance aggravates if active authors who are not true authors of the ``Target paper'' frequently appear with ``Target paper'':~\camel~will predict these active authors as the true authors because they appear frequently together with ``Target paper''.
% In consequence, if ``Target paper'' is written by an active author (i.e., an author with many publications), and if this author happens to appear frequently with ``Target paper'' in the same context window,~\camel~shows high accuracy. 
%active authors (i.e., authors with many publications) happen to appear frequently with ``Target paper'' in the same context window,~\camel~shows high accuracy. In other words, if active authors are true authors,~\camel~shows high accuracy.
%Revisiting \textbf{Toy Example} in Figure~\ref{fig:author}a in this perspective, Bob is more likely to be predicted as a true author of ``Target paper'' only if Bob is an active author.
%On the other hand, the model performs poorly if less active authors are true authors, and it aggravates if active authors who are not true authors of the ``Target paper'' frequently appear with ``Target paper''. 
While such behavior derived from the skip-gram model may be rational for general network embedding tasks whose goal is to preserve the overall proximity between nodes~\cite{dong2017metapath2vec,perozzi2014deepwalk,grover2016node2vec,fu2017hin2vec}, the skip-gram based objective should be reconsidered when it comes to a specific task, such as author identification.
We later show in our experiments that~\proposed~
is robust to the activeness of the authors, thanks to the pair validity classifier.
%by introducing pair embeddings together with the pair validity classifier.
%solves the above limitations by introducing pair embeddings together with a pair validity classifier.

There also exist several recent work that learn  relation type-specific node embeddings~\cite{shi2018easing,shi2018aspem}. However, the relation labels between nodes in their work are given in advance (e.g., a connection between two movies implies which genre they share), whereas relations between papers and authors in academic networks do not have predefined edge labels; we only know that a link exists between a paper and an author, if the paper is written by the author. Therefore, these methods cannot be directly applied to our setting.

%\subsection{Attention Mechanism}
\medskip
%\noindent\textbf{Attention Mechanism. }
%The attention mechanism has been firstly introduced for the task of sequence-to-sequence machine translation~\cite{bahdanau2014neural}, where a sequence of input is encoded and later decoded to another sequence of output.
%The main idea is to focus on the important part of the input data when decoding the output.
%There have been several network embedding methods that adopt the attention mechanism~\cite{qu2017attention,choi2017gram,velivckovic2017graph,lee2018graph}. However, none of them applied attention to model the context path between two nodes in a random walk.
%Other applications of the attention mechanism include image classification~\cite{xiao2015application}, recommendation~\cite{chen2017attentive} and speech recognition~\cite{chorowski2015attention}.

\section{Problem Definition}
In this section, we first introduce preliminary concepts regarding heterogeneous network, and then formalize the task to be addressed.
%author identification task to be addressed.

\begin{figure}[t]
	\centering
	\includegraphics[width=0.7\linewidth]{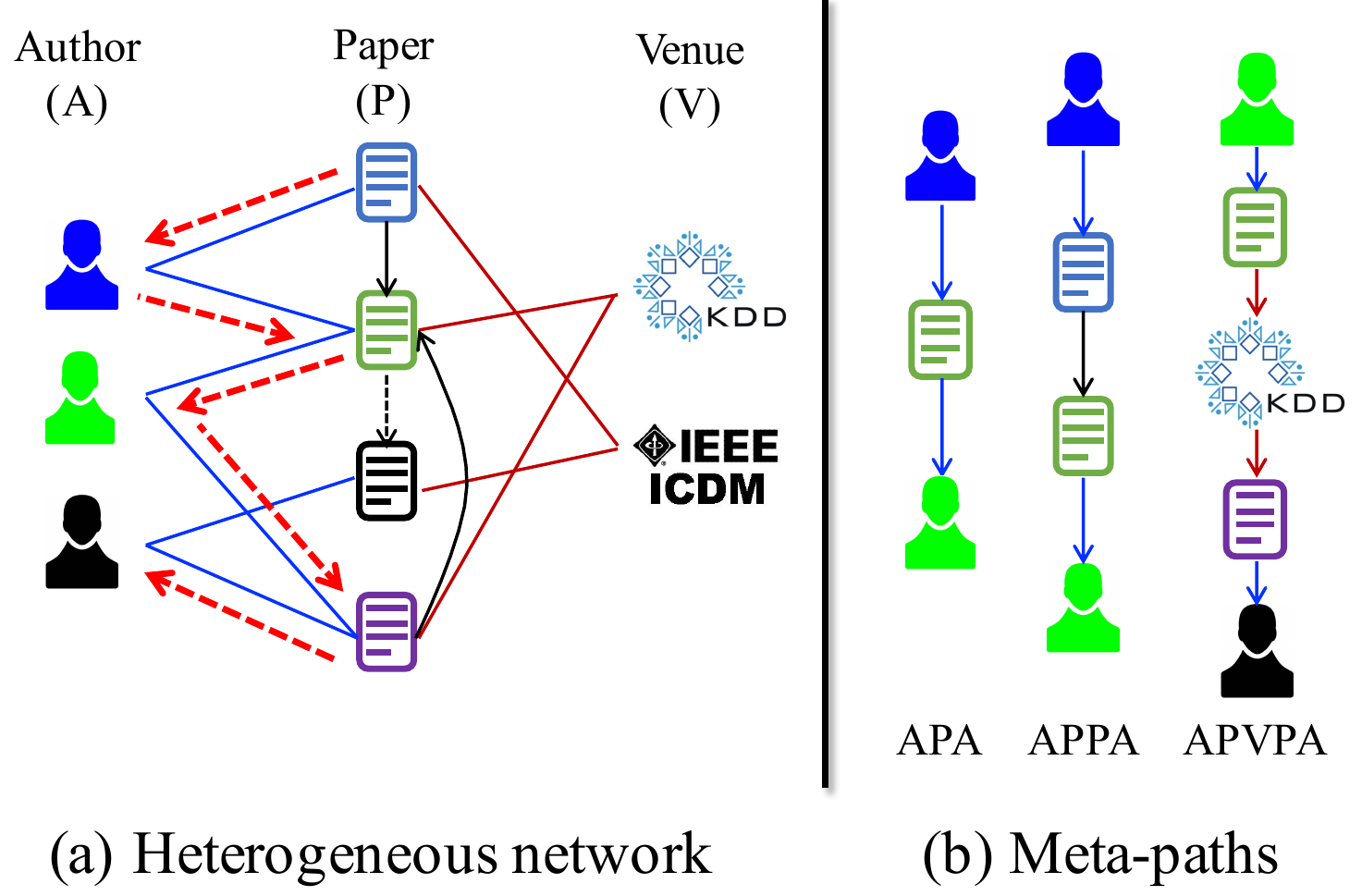}
%	\vspace{-2ex}	
	\caption{An illustrative example of (a) academic heterogeneous network. (b) Different meta-path schemes.}
	\vspace{-1ex}
	\label{fig:hetnet}
\end{figure}

\theoremstyle{definition}
\begin{definition}{(\textbf{Heterogeneous Network})}
	A heterogeneous network is a graph $\mathcal{G}=(\mathcal{V},\mathcal{E},\mathcal{T}_v,\mathcal{T}_e,\phi,\psi)$ in which $\mathcal{V}$ is the union of different types of nodes, $\mathcal{E}$ is the union of different types of edges. Each node $v\in\mathcal{V}$ and edge $e\in\mathcal{E}$ are associated with a node type mapping function $\phi: \mathcal{V} \rightarrow \mathcal{T}_v$, and an edge type mapping function $\psi: \mathcal{E} \rightarrow \mathcal{T}_e$, respectively.  $\mathcal{T}_v$ and $\mathcal{T}_e$ denote the sets of node types and edge types, respectively. $\mathcal{G}$ is defined as a heterogeneous network when the number of node types $|\mathcal{T}_v| > 1$  or the number of edge types $|\mathcal{T}_e| > 1$.
\end{definition}
\vspace{-0.5ex}
\begin{adjustwidth}{5pt}{5pt}
\noindent{\textbf{\underline{Example.} }}Figure~\ref{fig:hetnet}(a) shows an academic heterogeneous network that consists of three different types of nodes (Author (A), Paper (P), and Venue (V)), and three different types of edges (A$\leftrightarrow$P: author writes paper, P$\rightarrow $P: paper cites paper, and P$\leftrightarrow$V: paper publishes in venue).
\end{adjustwidth}

\begin{definition}{(\textbf{Meta-path})} Given $A_i\in\mathcal{T}_v$ and $R_i\in\mathcal{T}_e$, a meta-path~\cite{sun2011pathsim} $\mathcal{P}$  is defined as a path in the form of $A_1\xrightarrow{R_1}A_2\xrightarrow{R_2}\cdots \xrightarrow{R_{l-1}} A_{l}$ (abbreviated as $A_1A_2\cdots A_{l}$), which describes a composite relation $R=R_1\circ R_2 \circ \cdots \circ R_{l-1}$ between objects $A_1$ and $A_{l}$, where $\circ$ denotes the composition operator on relations. 
\end{definition}
\vspace{-0.5ex}
\begin{adjustwidth}{5pt}{5pt}
\noindent{\textbf{\underline{Example.} }}Figure~\ref{fig:hetnet}(b) shows three different ways that two authors can be connected: Author-Paper-Author (APA), Author-Paper-Paper-Author (APPA), and Author-Paper-Venue-Paper-Author (APVPA). Each meta-path contains different semantics. Specifically, APA means co-authorship, APPA means an author cites a paper written by another author, and APVPA means two authors publishing papers in the same venue.
\end{adjustwidth}

\begin{definition}{(\textbf{Meta-path guided Random Walk})} 
A meta-path guided random walk~\cite{dong2017metapath2vec} $\boldsymbol{w}\in \mathcal{W}^{\mathcal{P}}$ is a random walk guided by a specific meta-path $\mathcal{P}$, where $\mathcal{W}^{\mathcal{P}}$ is a set of collected random walks. Each random walk recursively samples a specific $\mathcal{P}$ until the walk length reaches a predefined value.
\end{definition}
\vspace{-0.5ex}
\begin{adjustwidth}{5pt}{5pt}
	\noindent{\textbf{\underline{Example.} }}Red dashed arrows in Figure~\ref{fig:hetnet}(a) shows a segment of a sample random walk guided by meta-path APA. In other words, a random walker in each step should only follow the pattern APA when deciding the next step. 
	
\end{adjustwidth}

\begin{figure}[t]
	\centering
	\includegraphics[width=0.8\linewidth]{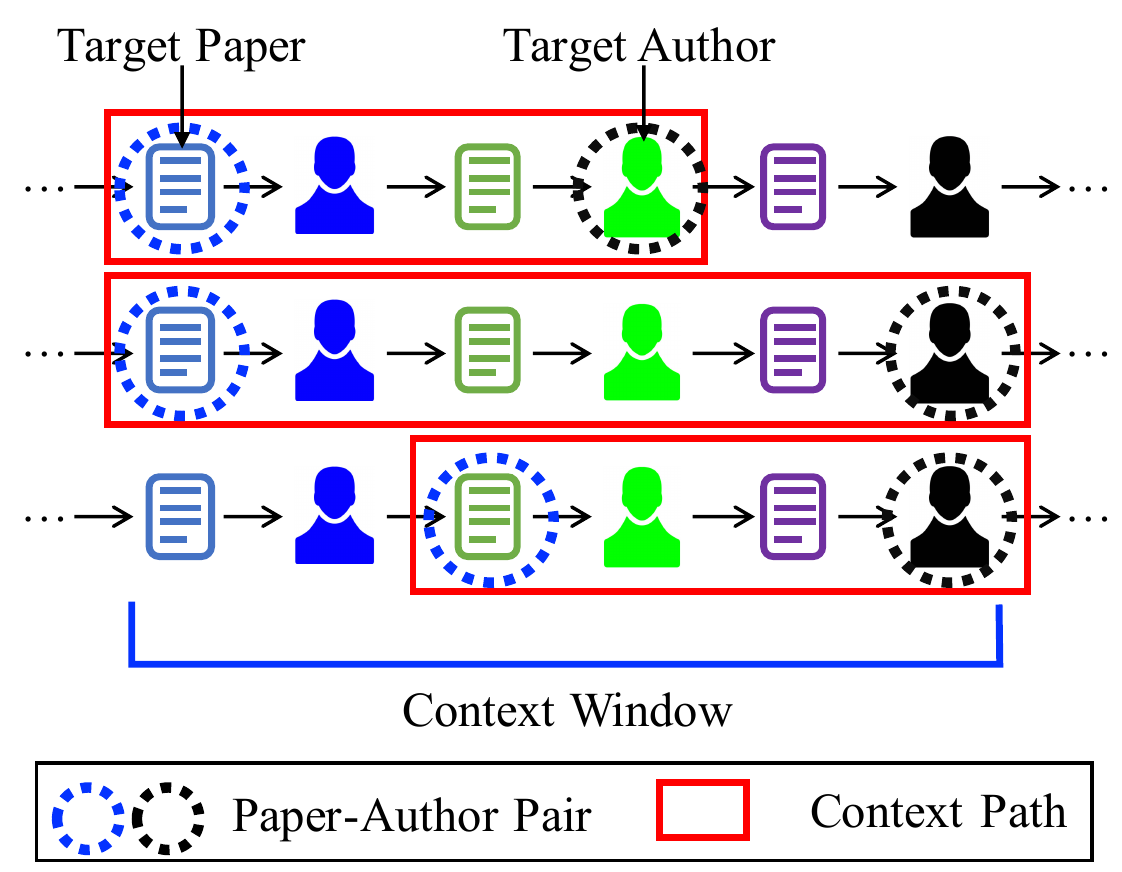}
	\caption{Examples of possible paper–author pairs, and their associated context paths within a segment of a meta-path guided random walk. }
	\label{fig:pair}
%	\vspace{-4ex}
\end{figure}

\begin{definition}{(\textbf{Context path})} 
Given two nodes $v_i,v_j \in \mathcal{V}$, a set of context paths from $v_i$ to $v_j$ for all walks $\boldsymbol{w}\in\mathcal{W}^\mathcal{P}$ are denoted by $\mathcal{C}^{\mathcal{P}}_{i\rightarrow j}$. 
\end{definition}
\begin{adjustwidth}{5pt}{5pt}
	\noindent{\textbf{\underline{Example.} }}Figure~\ref{fig:pair} shows an unfolded view of the sample random walk that follow the red dashed line shown in Figure~\ref{fig:hetnet}(a). Each red box denotes a context path of the associated paper--author pair in dashed circles.
\end{adjustwidth}

%\begin{table}[t]
%	\small
%	\begin{tabular}{ccc} \hline
%		Notation & \multicolumn{2}{l}{Description} \\ \hline
%		$\mathcal{P}$& \multicolumn{2}{p{6cm}}{\raggedright A meta-path.} \\
%		$\textbf{p}_v\in\mathbb{R}^K$& \multicolumn{2}{p{6cm}}{\raggedright Embedding vector of paper $v$.} \\
%		$\textbf{q}_u\in\mathbb{R}^K$& \multicolumn{2}{p{6cm}}{\raggedright Embedding vector of author $u$.} \\
%		$\textbf{Comb}(\cdot)$& \multicolumn{2}{p{6cm}}{\raggedright A function that combines $\textbf{p}_v$ and $\textbf{q}_u$.} \\
%		$\textbf{g}(\cdot)$& \multicolumn{2}{p{6cm}}{\raggedright Paper--author pair embedder ($\textbf{g}:\mathbb{R}^{4K} \rightarrow \mathbb{R}^d$).} \\
%		$\textbf{f}(\cdot)$& \multicolumn{2}{p{6cm}}{\raggedright Context path embedder ($\textbf{f}:\mathbb{R}^{K} \rightarrow \mathbb{R}^d$).}\\
%		$n$& \multicolumn{2}{p{6cm}}{\raggedright Number of layers of MLP in $\textbf{g}(\cdot)$.}\\
%		$\bm{\pi}(\cdot)$& \multicolumn{2}{p{6cm}}{\raggedright Pair validity classifier ($\bm{\pi}: \mathbb{R}^d \rightarrow \mathbb{R}$).}\\
%		$I_{<T}$& \multicolumn{2}{p{6cm}}{\raggedright A set of papers published before timestamp $T$.}\\			
%		$l_v$& \multicolumn{2}{p{6cm}}{\raggedright A set of true authors of paper $v$.}\\			
%		$\mathcal{W}^{\mathcal{P}}$& \multicolumn{2}{p{6cm}}{\raggedright A set of random walks guided by meta-path $\mathcal{P}$.}\\			
%		$\mathcal{S(\mathcal{P})}$& \multicolumn{2}{p{6cm}}{\raggedright A set of predefined meta-path schemes.}\\						
%		\hline
%	\end{tabular}
%\end{table}

\medskip
%\noindent With the aforementioned notations and definitions, the task to be addressed is formally defined as follows:
\noindent The task to be addressed is formally defined as follows:

%\noindent\textbf{Author Identification.} Given a set of papers published before timestamp $T$, each of which is associated 
%with bibliographic information, such as authors, references, abstract, published year and venue, our task is to rank all potential authors of each paper published after timestamp $T$, such that top-ranked authors are true authors.

\medskip
\begin{adjustwidth}{5pt}{5pt}
\noindent\textit{\textbf{Given}:} A set of node pairs $(v_i, v_j)$ and their associated set of context paths $\mathcal{C}^{\mathcal{P}}_{i\rightarrow j}$ extracted from multiple random walks guided by a meta-path $\mathcal{P}$ in a set of meta-path scheme $\mathcal{S}(\mathcal{P})$, 

\noindent\textit{\textbf{Goal}:} Predict the likelihood of the pairwise relationship between any two nodes in $\mathcal{V}$.
%$v_i$ and $v_j$.
%Predict the link probability between $v_i$ and $v_j$.
\end{adjustwidth}
\section{Heterogeneous Network Embedding}
\label{sec:het}
Inspired by skip-gram based word2vec~\cite{mikolov2013distributed,mikolov2018advances}, previous network representation learning methods~\cite{perozzi2014deepwalk,grover2016node2vec} viewed a network as a document. These methods first perform random walks on a network to extract multiple sequences of nodes, which are analogous to sequences of words, and then apply the skip-gram model to learn the representation of a node. 
%by viewing a sequence of nodes as a sentence. 
However, these methods only focus on homogeneous networks in which nodes and edges are of a single type. To learn effective representations of nodes in a heterogeneous network, metapath2vec~\cite{dong2017metapath2vec} introduced the meta-path guided random walk scheme together with the heterogeneous skip-gram model.

\smallskip 
\noindent\textbf{Meta-path Guided Random Walk. }
Given a heterogeneous network $\mathcal{G}=(\mathcal{V},\mathcal{E},\mathcal{T}_v,\mathcal{T}_e,\phi,\psi)$ and a meta-path $\mathcal{P}: A_1\xrightarrow{R_1}A_2\xrightarrow{R_2}\cdots A_i \xrightarrow{R_i} A_{i+1} \cdots \xrightarrow{R_{l-1}} A_{l}$, it first performs meta-path guided random walks to generate paths that capture both the semantic and structural correlations between multiple types of nodes. The transition probability of walk at step $t$ is defined as:

\begin{equation}
%\small
\label{eqn:metawalk}
p(v^{t+1}|v_i^t,\mathcal{P})=
\begin{cases}
\frac{1}{|N_{i+1}(v_i^t)|}, & (v^{t+1},v_i^t) \in \mathcal{E}, \psi(v^{t+1})=i+1 \\
0, & (v^{t+1},v_i^t) \in \mathcal{E}, \psi(v^{t+1})\neq i+1 \\
0, & (v^{t+1},v_i^t) \notin \mathcal{E}\\
\end{cases}
\end{equation}
where $v_i^t\in A_i$, and $N_{i+1}(v_i^t)$ denotes the $A_{i+1}$ type of neighborhood of node $v_i^t$. That is, the flow of a meta-path guided random walk is conditioned on the meta-path $\mathcal{P}$, and thus $v^{t+1}\in A_{i+1}$.

\smallskip 
\noindent\textbf{Heterogeneous Skip-gram Model. }
After generating a set of walks $\mathcal{W}^\mathcal{P}$ under meta-path $\mathcal{P}$ by performing the meta-path guided random walk as described above, metapath2vec learns the representation of nodes by maximizing the probability of having the heterogeneous context $N_{t}(v)$, $t\in\mathcal{T}_v$ given a node $v$:
\begin{equation}
%\small
\label{skipgram}
\operatorname*{arg\,max}_\theta \sum_{v \in \mathcal{V}} \sum_{t \in\mathcal{T}_v} \sum_{x_{t} \in N_{t}(v)} \log p\left(x_{t} | v ; \theta\right)
\end{equation}
where $x_t$ is a $t$-th type context node of $v$. The likelihood probability $p(x_t|v;\theta)$ is commonly defined as a softmax function, and Eqn.~\ref{skipgram} is optimized by adopting the negative sampling technique~\cite{mikolov2013distributed,dong2017metapath2vec}.

\section{The Pair Embedding Framework}
We present our novel task-guided pair embedding framework in heterogeneous network, called~\proposed.
As previously stated, we focus on the task of author identification as an application of our framework, because this task can be cast as predicting the likelihood of the pairwise relationship between a paper and an author. 
We first formally define the task of author identification as follows:
%our proposed method~\proposed~for solving the problem of author identification. 
%Then, we explain how to embed paper--author pairs and their associated context paths, which is followed by the pair validity classifier. Lastly, we present the final objective function of~\proposed. 

\medskip
\begin{adjustwidth}{5pt}{5pt}
\noindent\textbf{Task: Author Identification.} Given a set of papers published before timestamp $T$, each of which is associated 
with bibliographic information, such as authors, references, abstract, published year and venue, our task is to rank all potential authors of each paper published after timestamp $T$, such that top-ranked authors are true authors. 
\end{adjustwidth}

\subsection{Context Path-aware Pair Embedding}
To begin with, we perform multiple meta-path guided random walks as in Eqn.~\ref{eqn:metawalk} on the academic heterogeneous network, from which we extract paper--author pairs, and the associated context paths of each pair.
% as illustrated in Figure~\ref{fig:pair}.
Figure~\ref{fig:pair} shows how paper--author pairs and their associated context paths are extracted from a segment of a sample meta-path guided random walk. 
More precisely, given a paper--author pair in dashed circles, the associated context path is defined by a sequence of nodes between the paper and the author, including themselves. Note that we can generate three different pair-path instances from the random walk segment shown in Figure~\ref{fig:pair}.
In the following subsections, we will explain how the pairs and their context paths are embedded, respectively.
\subsubsection{\textbf{Embedding Paper--Author Pair}}
\label{sec:pair}
Given a combination of paper embedding $\textbf{p}_v\in\mathbb{R}^K$ and author embedding $\textbf{q}_u\in\mathbb{R}^K$ as $\textbf{Comb}(\textbf{p}_v,\textbf{q}_u)\in\mathbb{R}^{4K}$, the pair embedder $\textbf{g}:\mathbb{R}^{4K} \rightarrow \mathbb{R}^d$ is a multi-layer perceptron (MLP) with $n$ layers that generates a $d$-dimensional embedding for a paper--author pair $(v,u)$. 
%Precisely, we train a multi-layer perceptron (MLP) with $n$ layers on a combination of the paper and the author embeddings, i.e., $\textbf{Comb}(\textbf{p}_v,\textbf{q}_u)$, as follows:

%The pair embedder $\textbf{g}:\mathbb{R}^{4K} \rightarrow \mathbb{R}^d$ generates a $d$-dimensional embedding for a paper--author pair $(v,u)$. Precisely, we train a multi-layer perceptron (MLP) with $n$ layers on a combination of the paper embedding $\textbf{p}_v\in\mathbb{R}^K$ and the author embedding $\textbf{q}_u\in\mathbb{R}^K$, i.e., $\textbf{Comb}(\textbf{p}_v,\textbf{q}_u)$, as follows:
%\begin{equation}
%h^{(l)}=
%\begin{cases}
%\textsf{ReLU}(\textsf{LN}(W^{(l)}\textsf{Dropout}(h^{(l-1)}) + b^{(l)})), & 0<l < n\\
%\textsf{LN}(W^{(l)}\textsf{Dropout}(h^{(l-1)}) + b^{(l)}), & l=n\\
%\end{cases}
%\end{equation}
\begin{equation}
%\small
h^{(l)}=
\begin{cases}
\textsf{ReLU}(W^{(l)}\textsf{Dropout}(h^{(l-1)}) + b^{(l)}), & 0<l < n\\
W^{(l)}\textsf{Dropout}(h^{(l-1)}) + b^{(l)}, & l=n\\
\end{cases}
\end{equation}
\begin{equation}
\small
h^{(0)}=\textbf{Comb}(\textbf{p}_v,\textbf{q}_u)=[\textbf{p}_v;\textbf{q}_u;\\\textbf{p}_v\circ\textbf{q}_u;\textbf{p}_v-\textbf{q}_u]\in\mathbb{R}^{4K}
\end{equation}
\begin{equation}
\textbf{g}(v,u)=h^{(n)}\in\mathbb{R}^d
\end{equation}
where $\circ$ denotes element-wise vector multiplication, and $\textsf{ReLU}(x)=max(0,x)$.
We apply dropout~\cite{srivastava2014dropout} on the hidden layers, and take the last layer output of MLP as the pair embedding. In the experiments, $\textbf{g}(\cdot)$ is a 2-layered MLP each layer with 100 hidden units.
%, and adopt layer normalization $\textsf{LN}(\cdot)$ before each activation layer to normalize neuron activations across layer~\cite{lei2016layer}. Our experiments show that $\textsf{LN}(\cdot)$ helps~\proposed~converge fast.

For author identification, 
%as the target papers of which we aim to predict authors are not known during the training process, 
we need to represent a paper using information related to the paper.
%by trained parameters.
%using parameters that are used during training. 
HNE~\cite{chen2017task} represents a paper by combining the embeddings of its references, keywords, and venue.
%, and keywords, which are combined to represent papers used during training.
However, as the semantic content of a paper is critical for identifying the authors of the paper,~\camel~uses word embeddings~\cite{mikolov2013distributed} to represent a paper by its content~\cite{zhang2018camel}, i.e., abstract, which considerably outperformed HNE. 
Hence, we also use the words in the abstract of a paper to represent the paper:
\begin{equation}
\label{eqn:paperencoder}
\textbf{p}_{v}=\textsf{PaperEncoder}\left({O}_{v}\right)\in\mathbb{R}^K
\end{equation}
where $O_v$ is the index of paper $v$, and $\textsf{PaperEncoder}(\cdot)$ is a GRU--based content encoder, which encodes a paper into a vector. 
We adopt $\mathcal{L}_{Metric}$ of~\camel~\cite{zhang2018camel} to encode the paper content. 
Please refer to the original paper for more details~\cite{zhang2018camel}.

\subsubsection{\textbf{Embedding Context Path}}
\label{sec:path}
\begin{figure}[t]
	\centering
	\includegraphics[width=0.9\linewidth]{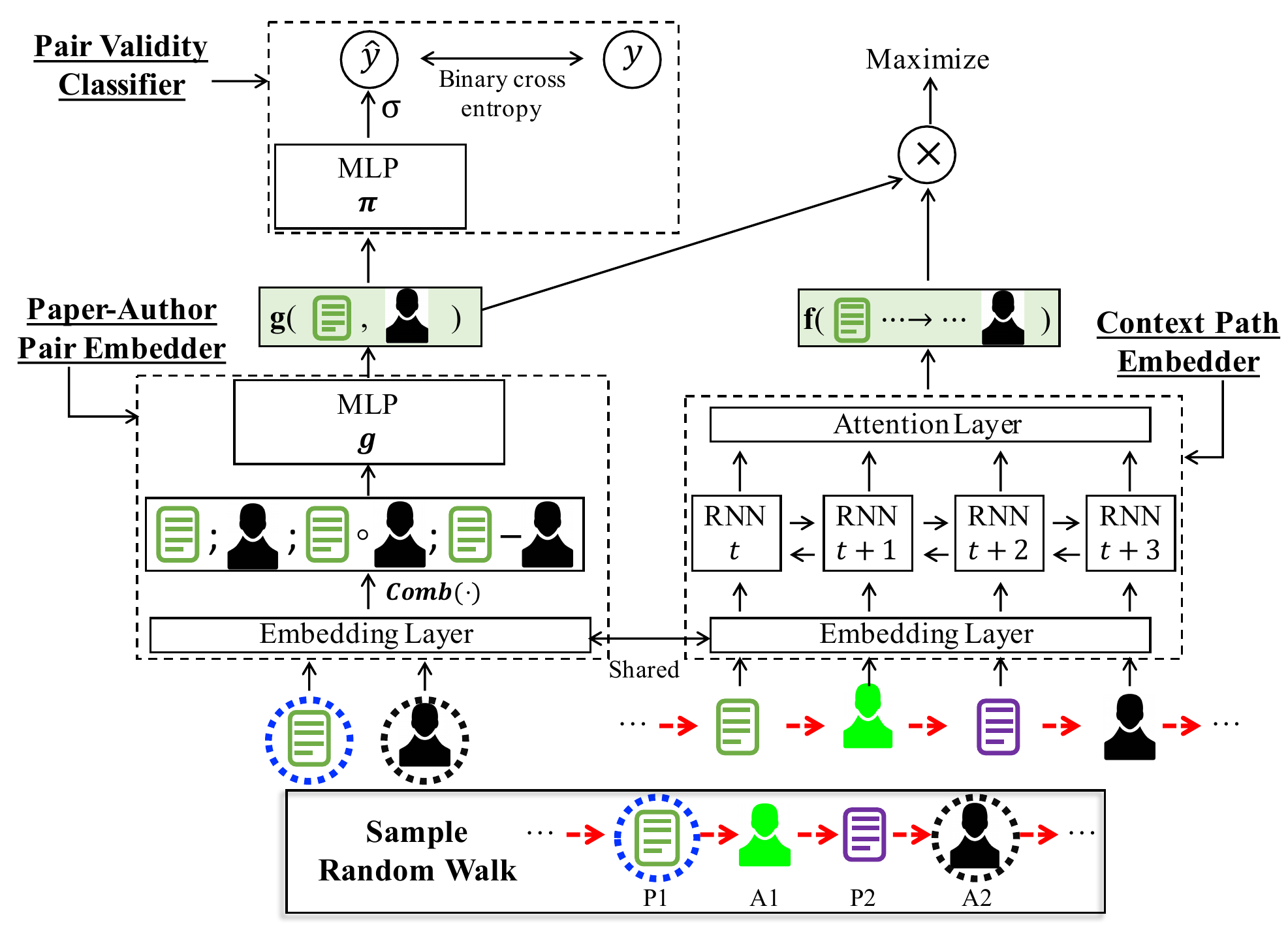}
	\caption{An overview of~\proposed.}
	\label{fig:overview}
		\vspace{-2ex}
\end{figure}

Recall from Figure~\ref{fig:pair} that given a paper--author pair $(v,u)$ on a meta-path guided random walk guided by $\mathcal{P}$, there exists a set of context paths $\mathcal{C}^{\mathcal{P}}_{v\rightarrow u}$, i.e., a set of node sequences between paper $v$ and author $u$.
Our assumption is that we can readily \textit{infer the research topic related to the pair $(v,u)$ by examining the path between paper $v$ and author $u$}. For example, in the sample random walk shown in~Figure~\ref{fig:overview}, P1 and P2 are likely to be about a similar research topic because both of them are written by A1. Besides, A1 and A2 are likely to share a common research interest because they co-authored P2. In short, the context path from P1 to A2 reveals the research topic related to the pair (P1, A2).

Given a  set of paper--author pairs and their associated context paths obtained from meta-path guided random walk~\cite{zhang2018camel,dong2017metapath2vec}, we embed a context path composed of a sequence of nodes by applying bidirectional gated recurrent unit (GRU)~\cite{cho2014learning}, which is commonly used for sequence modeling. 
%Assuming that we have predefined meta-paths $\mathcal{P}$,
%we perform meta-path guided random walk~\cite{zhang2018camel,dong2017metapath2vec} as in Eqn.~\ref{eqn:metawalk}, and then obtain a set of paper--author pairs and their context paths within the walks.
%In order to embed a context path composed of a sequence of nodes, we apply bidirectional gated recurrent unit (GRU)~\cite{cho2014learning}, which is commonly used for sequence modeling. 
%followed by attention module. 
More precisely, the context path embedder $\textbf{f}:\mathbb{R}^{K} \rightarrow \mathbb{R}^d$ is a bidirectional GRU that generates a $d$-dimensional vector for each context path by adopting an attention module. Take a context path between paper $v$ and author $u$, i.e., ${c}\in\mathcal{C}^{\mathcal{P}}_{v\rightarrow u}$, as an example, which is represented as a sequence of nodes, i.e.,  ${c}=\{v,c_2,c_3,...,c_{n-1},u\}$. We convert this sequence of nodes into a sequence of $K$-dimensional embedding vectors $\{\textbf{p}_v, \textbf{x}_2, \textbf{x}_3, ...,\textbf{x}_{n-1}, \textbf{q}_u\}$, where $\textbf{p}_v=\textbf{x}_1$ and $\textbf{q}_u=\textbf{x}_n$.
Note that the types of nodes $c_2,c_3...,c_{n-1}$ depend on what kind of meta-path $\mathcal{P}$ guided the random walk. For example, if $\mathcal{P}=${Author $\rightarrow$ Paper $\rightarrow$ Author}, then $\phi(c_2)=$ Author, $\phi(c_3)=$ Paper, and $\phi(c_{n-1})=$ Paper.
At time $t$, GRU computes the hidden state $\textbf{h}_t\in\mathbb{R}^d$ given the previous hidden state $\textbf{h}_{t-1}\in\mathbb{R}^d$ and the current input $\textbf{x}_t\in\mathbb{R}^K$, i.e., $\textbf{h}_t=\textsf{GRU}(\textbf{h}_{t-1},\textbf{x}_t)$. More precisely, 
\begin{equation}
\small
\begin{array}{l}
{\textbf{z}_{t}=\sigma\left(\textbf{W}_{z} \textbf{x}_{t}+\textbf{U}_{z} \textbf{h}_{t-1} + \textbf{b}_z\right)},\
 {\textbf{r}_{t}=\sigma\left(\textbf{W}_{r} \textbf{x}_{t}+\textbf{U}_{r} \textbf{h}_{t-1}+\textbf{b}_r\right)} \\
 {\hat{\textbf{h}}_{t}=\tanh \left[\textbf{W}_{h} \textbf{x}_{t}+\textbf{U}_{h}\left(\textbf{r}_{t} \circ \textbf{h}_{t-1}\right)+\textbf{b}_h\right]} \\ {\textbf{h}_{t}=\textbf{z}_{t} \circ \textbf{h}_{t-1}+\left(1-\textbf{z}_{t}\right) \circ \hat{\textbf{h}}_{t}}
 \end{array}
\end{equation}
where $\sigma(\cdot)$ is a sigmoid function, $\textbf{W}\in\mathbb{R}^{d\times K}$, $\textbf{U}\in\mathbb{R}^{d\times d}$ and $\textbf{b}\in\mathbb{R}^{d}$ are parameters of GRU network.
In order to make full use of the context information from both directions, we apply bidirectional GRU, i.e., $\textbf{h}_t=\textsf{BiGRU}(\overrightarrow{\textbf{h}}_{t-1},\overleftarrow{\textbf{h}}_{t-1},\textbf{x}_t)$ and combine the output from each direction by a linear projection layer. More precisely,
\begin{equation}
\small
\label{eqn:ctxrnn}
\begin{array}{l}
\overrightarrow{\textbf{h}}_t=\textsf{GRU}(\overrightarrow{\textbf{h}}_{t-1},\textbf{x}_t),\ \overleftarrow{\textbf{h}}_t=\textsf{GRU}(\overleftarrow{\textbf{h}}_{t-1},\textbf{x}_t) \\
\textbf{h}_t=\textbf{W}_{\text{proj}}\text{[}\overrightarrow{\textbf{h}}_t ; \overleftarrow{\textbf{h}}_t\text{]} + \textbf{b}_{\text{proj}}
\end{array}
\end{equation}
where $\textbf{W}_{\text{proj}}\in\mathbb{R}^{d\times 2d}$ and $\textbf{b}_{\text{proj}}\in\mathbb{R}^{d}$ are parameters of the linear projection layer, and $[\cdot;\cdot]$ denotes vector concatenation operator. 

After embedding $n$ nodes within a context path ${c}\in\mathcal{C}^{\mathcal{P}}_{v\rightarrow u}$ into an embedding matrix $\textbf{h}\in\mathbb{R}^{n\times d}$, we aggregate the matrix $\textbf{h}$ by applying attentive pooling~\cite{zhou2016attention} that extracts a $d$-dimensional vector from $\textbf{h}$, which summarizes the context path by measuring the contribution of each node in the context path to form a high-level representation of the entire context path. More precisely, the path embedder $\textbf{f}$ is defined as follows:
\begin{equation}
%\small
\begin{aligned}
& w_{t} =\text{softmax}(\textbf{kh}_t)=\frac{\text{exp}({\mathrm{\textbf{k}} \textbf{h}_{\mathrm{t}})}}{\sum_{i=1}^{n} \text{exp}({\mathrm{\textbf{kh}}_{\mathrm{i}})}},\ \textbf{f}(c) =\sum_{t} w_{t} \mathrm{\textbf{W}_{\text{attn}}\textbf{h}}_{t}
\end{aligned}
\label{eqn:attn}
\end{equation}

%\begin{equation}
%\textbf{f}(c) =\sum_{t} w_{t} \mathrm{\textbf{W}_{\text{attn}}\textbf{h}}_{t}
%\end{equation}
\noindent where $\textbf{k}\in\mathbb{R}^d$ and $\textbf{W}_{\text{attn}}\in\mathbb{R}^{d\times d}$. The attention module enables us to pay attention to more important segments in a context path. 
Note that we have also tried other types of attention~\cite{luong2015effective}, but attentive pooling performed the best. We conjecture that as attentive pooling is specifically designed for relation classification tasks~\cite{zhou2016attention}, it is suitable for author identification in which identifying the paper--author relationship is critical. 
In short, we expect $\textbf{f}(c)\in\mathbb{R}^d$ to encode the research topic associated with the context path $c$.

\subsubsection{\textbf{Injecting Context Information into Pairs}}
\label{sec:joint}
Given a paper--author pair $(v,u)$, our goal is to minimize the negative log likelihood of $(v,u)$ given a context path ${c}\in\mathcal{C}^{\mathcal{P}}_{v\rightarrow u}$:
\begin{equation}
%\small
\label{eqn:l1}
\mathcal{L}_{\text{ctx}}(v,u)= \sum_{{c}\in\mathcal{C}^{\mathcal{P}}_{v\rightarrow u}}
- \log p((v,u)|c,\mathcal{P})
\end{equation}
The likelihood probability $ p((v,u)|c,\mathcal{P})$ is defined as follows:
\begin{equation}
%\small
\label{eqn:softmax}
p((v,u)|c,\mathcal{P})=\frac{\exp \left[\left(\textbf{g}(v, u) \cdot \textbf{f}(c)\right) \right]}{\sum_{c^{\prime} \in C_*^{\mathcal{P}}} \exp \left[\left(\textbf{g}(v, u) \cdot \textbf{f}(c^\prime)\right)\right]}
\end{equation}
where $C_*^{\mathcal{P}}$ denotes all possible context paths that can be obtained from $\mathcal{W}_\mathcal{P}$. As directly optimizing Eqn.~\ref{eqn:softmax} is computationally expensive owing to a large number of possible context paths $C_*^{\mathcal{P}}$, we apply commonly used negative sampling technique~\cite{mikolov2013distributed,zhang2018camel,dong2017metapath2vec} for optimization. More precisely, Eqn.~\ref{eqn:softmax} can be approximated as:
\begin{equation}
\small
\label{eqn:l1_}
\log p((v,u)|c,\mathcal{P})\approx\log \sigma(\textbf{g}(v, u) \cdot \textbf{f}(c))+\sum_{j=1}^{k} \log \sigma\left(-\textbf{g}(v, u) \cdot \textbf{f}\left(c_{\text{rand}}^{j}\right)\right)
\end{equation}
where $k$ is the number of randomly sampled contexts for pair $(v,u)$. 
%After applying negative sampling technique to randomly generate context paths for each pair $(v,u)$, 
That is, we wish to make $\textbf{g}(v,u)$ similar to $\textbf{f}(c)$, if $c$ is a context path between $v$ and $u$, and keep $\textbf{g}(v,u)$ dissimilar to the embeddings of random context paths $\textbf{f}\left(c_{\text{rand}}\right)$. 
By maximizing Eqn.~\ref{eqn:l1}, the pair embedding $\textbf{g}(v,u)$ will naturally get similar to the embeddings of more frequently appearing context paths. This in turn facilitates $\textbf{g}(v,u)$ to encode its related research topic, because frequently appearing paths define the relationship between the pair.

Note that Eqn.~\ref{eqn:l1} has a similar underlying philosophy as the heterogeneous skip-gram model in Eqn.~\ref{skipgram}, which is to predict context nodes given the center node. However, the difference of our proposed method compared with skip-gram model lies in the perspective that a center node is extended to a pair of nodes, and a context node is extended to a context path. 
\subsection{Validity of Pair Embedding}
\label{sec:validity}
Recall that one of the limitations of~\camel~was that it 
%blindly trains two nodes to be similar to each other if they appear within the same context window, which
% inadvertently makes a paper similar to author
% favors the pairs 
%that frequently appear together within a context window of a random walk sequence. 
inadvertently makes a paper embedding and an author embedding similar to each other
if they frequently appear together within the same context window of a random walk sequence, whether or not the author is a true author of the paper. 
Therefore,~\camel~shows high accuracy when the true authors are active authors with many papers, but performs poorly when the true authors are relatively inactive authors, which is an inevitable consequence of the skip-gram model (refer to our discussions on Figure~\ref{fig:camel_walk}). 
%an active author (i.e., an author with many publications) is more likely to appear frequently together with ``Target paper'' within the same context window than an inactive author
%the frequently appearing authors of a paper are in fact not the true authors the paper, which is an inevitable consequence of the skip-gram model. 
To this end, we propose a pair validity classifier $\bm{\pi}: \mathbb{R}^d \rightarrow \mathbb{R}$ to discriminate whether the paper--author pair is a valid pair or not, which is formulated by binary cross-entropy loss as follows:
\begin{equation}
%\small
\mathcal{L}_{\text{pv}}(v,u)= y_{v,u} \sigma(\bm{\pi}(\textbf{g}(v, u)))+(1-y_{v,u}) (1-\sigma(\bm{\pi}(\textbf{g}(v, u))))
\end{equation}
\begin{equation}
%\small
y_{v,u}=
\begin{cases}
1, & \text{paper $v$ is written by author $u$} \\
0, & \text{paper $v$ is not written by author $u$} \\
\end{cases}
\end{equation}
%As the validity classifier injects the pair validity information into the pair embedding, 
$\bm{\pi}(\cdot)$ is a 2-layered MLP with ReLU activation.
Armed with the pair validity classifier that explicitly discriminates the validity of each pair with respect to the specific task at hand,~\proposed~is able to not only identify active authors with many publications, but also relatively less active authors, because the training of the embedding vectors is no longer solely based on the frequency.
Moreover, thanks to the pair validity classifier, two nodes that constitute a pair will be close to each other not only if they are related to a similar research topic, but also the pair itself is valid at the same time.
%pairs will be embedded similar to each other not only when they share similar research interest, but also when they are valid pairs at the same time. 
%In other words, pairs will be embedded close if they are similar in terms of research interest, and if they are valid.

\subsection{Joint Objective}
Combining $\mathcal{L}_\text{ctx}(v,u)$ and $\mathcal{L}_\text{pv}(v,u)$ for all possible $(v,u)$ pairs in meta-path guided random walks guided by $\mathcal{P}$, we obtain the following objective function $\mathcal{L}$:
\begin{equation}
%\small
\label{eqn:final}
\mathcal{L} =\sum_{\mathcal{P} \in \mathcal{S(\mathcal{P})}}\sum_{\boldsymbol{w} \in \mathcal{W}_{\mathcal{P}}} \sum_{v \in \boldsymbol{w}} \sum_{u \in \boldsymbol{w}\left[C_{v}-\tau : C_{v}+\tau\right]} \left[\mathcal{L}_{\text{ctx}}(v,u) + \mathcal{L}_{\text{pv}}(v,u)\right]
\end{equation}
where $\mathcal{S(\mathcal{P})}$ denotes all predefined meta-path schemes, $\tau$ is the context window size of paper $v$, and $C_v$ denotes the position of $v$ in walk $w$. The final objective function $\mathcal{L}$ can be minimized by using mini-batch Adam optimizer~\cite{kingma2014adam}.
Figure~\ref{fig:overview} illustrates the overall model architecture of~\proposed.

\medskip
\noindent{\textbf{Prediction. }}
The final prediction of~\proposed~is computed by the output of the pair validity classifier $\bm{\pi}(\cdot)$. Precisely, if we were to identify true authors of paper $v$, we rank $\sigma(\bm{\pi}(\textbf{g}(v, u)))$ for $u\in U$, where $U$ denotes the set of users, to see how many top-ranked authors are true authors. We only need to know the abstract content of a paper for identifying its true authors, because $\textsf{PaperEncoder}(\cdot)$ and author embeddings are learned during training.

\begin{table*}[t]
	\centering
	\small
	\caption{The overall performance on author identification (Impr. denotes improvements of~\proposed~over the best  baseline).}
	\vspace{-2ex}
	\renewcommand{\arraystretch}{0.5}
	\begin{tabular}{c|c|>{\arraybackslash}p{0.85cm}|>{\centering\arraybackslash}p{0.5cm}>{\centering\arraybackslash}p{0.5cm}>{\centering\arraybackslash}p{0.5cm}>{\centering\arraybackslash}p{0.8cm}|>{\centering\arraybackslash}p{0.9cm}c|c||c|>{\centering\arraybackslash}p{0.5cm}>{\centering\arraybackslash}p{0.5cm}>{\centering\arraybackslash}p{0.5cm}>{\centering\arraybackslash}p{0.8cm}|>{\centering\arraybackslash}p{0.9cm}c|c}
		\multicolumn{2}{c|}{Dataset} & \multicolumn{1}{c|}{Metric} & Sup  & MPV   & HNE   & \camel & \proposed$_{\text{npv}}$ & \proposed & Impr. &  & Sup  & MPV   & HNE   & \camel &\proposed$_{\text{npv}}$ & \proposed & Impr. \\
		\midrule
		\multirow{18}[1]{*}{\rotatebox[origin=c]{90}{AMiner-Top}} & \multirow{8}[1]{*}{\rotatebox[origin=c]{90}{$T$=2013}} & Rec@5 & 0.5460 &   0.5274    & 0.4874 & 0.5902& 0.6405& \textbf{0.6807} & 15.33\% & \multirow{18}[1]{*}{\rotatebox[origin=c]{90}{AMiner-All}} &   0.6096    &    0.5990   &    0.6110   &   0.5458    & 0.7049& \textbf{0.7097}    &  16.15\%\\
		&       & Rec@10 & 0.6227 &  0.6746     & 0.6301 & 0.7370 &0.7677 & \textbf{0.7849} & 6.50\% &      &    0.6409    &   0.7317    &   0.7166    &   0.6811    &0.8121  & \textbf{0.8237}    &12.57\%   \\
		&       & Prec@5 & 0.2285 &   0.2148    & 0.2051 & 0.2439 &0.2662 & \textbf{0.2835} & 16.24\% &      &  0.2679     &   0.2562    &   0.2679    &   0.2393    & 0.3076&  \textbf{0.3087}  & 15.23\% \\
		&       & Prec@10 & 0.1323 &   0.1401    & 0.1334 & 0.1555 &0.1632 &\textbf{0.1664} & 7.01\% &       &    0.1418   &   0.1595    &   0.1590    &   0.1508  & 0.1795&\textbf{0.1818}    & 13.98\% \\
		&       & F1@5  & 0.3222 &   0.3052    & 0.2888 & 0.3452 &0.3761& \textbf{0.4003} & 15.96\% &       &    0.3722   &   0.3589    &    0.3724   &    0.3327 & 0.4283& \textbf{0.4303}      &15.55\%  \\
		&       & F1@10 & 0.2182 &   0.2320    & 0.2202 & 0.2568& 0.2691 &\textbf{0.2746} & 6.93\% &       &   0.2322    &   0.2619    &    0.2602   &   0.2470  & 0.2940 &\textbf{0.2978}     & 13.71\% \\
		&       & AUC   & 0.7817 &    0.8887   & 0.8614 & 0.9112 &0.9164 &\textbf{0.9178} & 0.72\% &       &     0.7641  &   0.8923    &    0.8855   &    0.8768 & 0.9291 & \textbf{0.9337}    & 4.64\% \\
		\cmidrule{2-10}
		\cmidrule{12-18}
		& \multirow{9}[0]{*}{\rotatebox[origin=c]{90}{$T$=2014}} 
		& Rec@5 & 0.5142 & 0.5116      & 0.4665 & 0.5625& 0.6121 & \textbf{0.6577} & 16.92\% &       &    0.6203   &    0.5768   &   0.5842    &   0.5494   			&0.6742&   \textbf{0.6840} &  10.27\%\\
		&       & Rec@10 & 0.5792 &   0.6661    & 0.6185 & 0.7198 & 0.7471 & \textbf{0.7698} & 6.95\% &       &  0.6570     &   0.7114    &   0.6927    &   0.6835   	&0.7952	&   \textbf{0.7998} & 12.43\% \\
		&       & Prec@5 & 0.2508 &     0.2457  & 0.2284 & 0.2706 & 0.2962 & \textbf{0.3148} & 16.33\% &       &   0.2825    &  0.2586     &    0.2689   &    0.2529  & 0.3068&   \textbf{0.3109} &  10.05\%\\
		&       & Prec@10 & 0.1447 &  0.1636     & 0.1538 & 0.1776 & 0.1851 &\textbf{0.1898} & 6.87\% &       &   0.1510    &   0.1623    &    0.1611   &    0.1588   	  &  0.1840 &  \textbf{0.1850}  &  13.99\%\\
		&       & F1@5  & 0.3371 &   0.3320    & 0.3066 & 0.3654 & 0.3992 &\textbf{0.4258} & 16.53\% &       &   0.3882    &      0.3571 &   0.3683    &   0.3464    &0.4217&   \textbf{0.4275}	   & 10.12\% \\
		&       & F1@10 & 0.2316 &     0.2627  & 0.2463 & 0.2849 &0.2967 &\textbf{0.3045} & 6.88\% &       &   0.2455    &   0.2643    &   0.2614    &   0.2577   		& 0.2989 & \textbf{0.3005}    &  13.70\%\\
		&       & AUC   & 0.7359 &   0.8904    & 0.8619 & 0.9087 & 0.9112 &\textbf{0.9206} & 1.31\% &       &    0.7829   &   0.8834    &    0.8747   &  0.8770     		&0.9243 &  \textbf{0.9245} &  4.65\%\\
	\end{tabular}%
	\label{tab:overall}%
%	\vspace{-3ex}
\end{table*}%

\section{Experiments}
\label{sec:experiment}
The experiments are designed to answer the following research questions (\textbf{RQs}):
\begin{itemize}
	\item [\textbf{RQ 1}] How does~\proposed~perform compared with other state-of-the-art methods?
	\item [\textbf{RQ 2}]  How does~\proposed~perform on less active authors (i.e., users with few publications)?
	\begin{itemize}
		\item Qualitative analysis (\textbf{RQ 2-1})
	\end{itemize}
%	\item [\textbf{RQ 2}] How does~\proposed~perform on less active authors (i.e., users with few publications)?
%	\item [\textbf{RQ 3}] Does~\proposed~solve the limitation of skip-gram based model mentioned in Figure~\ref{fig:camel_walk}?
%	 from which~\camel~suffer?
	\item [\textbf{RQ 3}] How does each component of~\proposed~contribute to the overall performance (Ablation study)?
	\item [\textbf{RQ 4}] How are the pair embeddings visualized compared with paper/author embeddings?
%	that inadvertently makes a paper embedding and an author embedding
%similar to each other if they frequently appear together within a
%context window of a random walk sequence, whether or not the
%author is a true author of the paper.
\end{itemize}

\subsection{Experimental Setup}
\noindent\textbf{Dataset. }
In order to make fair comparisons with~\camel~\cite{zhang2018camel}, we evaluate our proposed method on AMiner dataset\footnote{https://aminer.org/citation}, which is an academic collaboration platform in computer science domain. As the preprocessed dataset is not available, we preprocess the data to make similar statistics with~the dataset used in~\cite{zhang2018camel}. More precisely, we extracted 10 years of data from 2006 to 2015, removed the papers published in venues with limited publications (e.g., workshop or tutorial)  and papers without abstract text. Moreover, as most researchers pay attention to top venues, and their research areas can be categorized into several different areas, we additionally generate a subset data of six research areas (AMiner-Top) according to Google Scholar Metrics: Artificial Intelligence (AI), Data Mining (DM), Databases (DB), Information System (IS), Computer Vision (CV) and Computational Linguistics (CL). For each research area, we choose three top venues that are considered to have influential papers~\footnote{\textbf{AI}: ICML, AAAI, IJCAI. \textbf{DM}: KDD, WSDM, ICDM. \textbf{DB}: SIGMOD, VLDB, ICDE. \textbf{IS}: WWW, SIGIR, CIKM. \textbf{CV}: CVPR, ICCV, ECCV. \textbf{CL}: ACL, EMNLP, NAACL}. 
In the end, AMiner-Top dataset contains 27,920 authors, 21,808 papers and 18 venues, and AMiner-Full dataset contains 536,811 authors, 447,289 papers and 389 venues.
%We summarize the statistics of the two datasets used in our experiments (AMiner-Top and AMiner-Full) in Table~\ref{tab:stats}.

\medskip
\noindent\textbf{Methods Compared.}
As~\proposed~is a task-guided heterogeneous network embedding framework, and we focus on the problem of author identification, we choose the following baselines.
\begin{enumerate}[leftmargin=0.2in]
	\item Feature engineering--based supervised method.
	\begin{itemize}[leftmargin=0.1in]
		\item \textbf{Sup}: Triggered by KDD Cup 2013, the problem of author identification has recently garnered attention, and top solutions of the challenge heavily relied on feature engineering followed by supervised ranking models on these features~\cite{efimov2013kdd,li2015combination}. Following them, we extract 16 features for each pair of paper and author in the training set. For more details about the features, refer to Table 2 of Zhang et al.,~\cite{zhang2018camel}. As for the ranking model, we tried logistic regression, support vector machine, gradient boosting, random forest and multi-layer neural network (NeuN), and found that NeuN performed the best.
	\end{itemize}
	\item General purpose heterogeneous network embedding method.
	\begin{itemize}
		\item \textbf{metapath2vec++ (MPV)~\cite{dong2017metapath2vec}}: The state-of-the-art heterogeneous network embedding method based on meta-path guided random walk that learns general node embeddings. To directly compare within our setting, where papers in test data are not known during training, we represent a paper by the words of its abstract. We adopt a GRU--based $\textsf{PaperEncoder}(\cdot)$ shown in Eqn.~\ref{eqn:paperencoder} to encode a paper into a vector.
	\end{itemize}
	\item Task-guided heterogeneous network embedding methods.
	\begin{itemize}
		\item \textbf{HNE~\cite{chen2017task}}: A task-guided heterogeneous network embedding method that resort to the network structure of an academic network rather than exploring the paper content.
		\item \textbf{\camel~\cite{zhang2018camel}}: The state-of-the-art heterogeneous network embedding--based method for author identification in which task-dependent and content-aware skip-gram model is proposed to formulate the correlations between each paper	and its indirect author neighbors.
		\item \textbf{\proposed$_{\text{npv}}$}: A variant of~\proposed~in which instead of the pair validity classifier, a dot product is used. i.e., $\mathcal{L}_{\text{pv}}(v,u)= y_{v,u} \sigma(\textbf{p}_v^T\textbf{q}_u)+(1-y_{v,u}) (1-\sigma(\textbf{p}_v^T\textbf{q}_u))$.
%		An ablation of~\proposed~in which instead of scoring with $\sigma(\bm{\pi}(\textbf{g}(v, u)))$, dot product of paper and author embeddings are used for scoring as done in~\camel.
%		instead of pair embeddings, paper--author pair embeddings are used for prediction as done in~\camel.
	\end{itemize}
\end{enumerate}
We do not compare with homogeneous network embedding methods, such as, DeepWalk~\cite{perozzi2014deepwalk} and node2vec~\cite{grover2016node2vec}, as their performance have been surpassed by methods designed for heterogeneous network embedding~\cite{dong2017metapath2vec}.

\medskip
\noindent\textbf{Evaluation Metrics. }
Recall that our task is to rank candidate authors for each paper $v\in I_{\geq T}$ in test dataset, where $I_{\geq T}$ denotes papers published after timestamp $T$. Hence, we use four commonly used ranking metrics, i.e., Recall@N, Precision@N, F1 score and AUC, to evaluate the performance of each method. 
Recall@N measures the ratio of the number of true authors among top-N predicted ranked list over the total number of true authors, Precision@N measures the ratio of the number of true authors in top-N predicted ranked list, and F1 measures the harmonic mean of precision and recall: F1 is high only if both precision and recall are high.
AUC measures the probability of a positive instance being ranked higher than a randomly chosen negative one.

\begin{table*}[t]
	\centering
	\small
	\caption{Author identification performance on relatively inactive users (\#papers $\leq$ 5) (vs.~\camel). }
	\vspace{-2ex}
	\renewcommand{\arraystretch}{0.8}
	\begin{tabular}{c|>{\centering\arraybackslash}m{0.5cm}|>{\centering\arraybackslash}m{1.3cm}|>{\centering\arraybackslash}m{0.8cm}>{\centering\arraybackslash}m{0.8cm}>{\centering\arraybackslash}m{0.8cm}>{\centering\arraybackslash}m{0.8cm}|>{\centering\arraybackslash}m{0.8cm}>{\centering\arraybackslash}m{0.8cm}>{\centering\arraybackslash}m{0.8cm}>{\centering\arraybackslash}m{0.8cm}|>{\centering\arraybackslash}m{0.8cm}>{\centering\arraybackslash}m{0.8cm}>{\centering\arraybackslash}m{0.8cm}>{\centering\arraybackslash}m{0.8cm}|>{\centering\arraybackslash}m{0.8cm}}
		\toprule
		\multirow{2}[1]{*}{\rotatebox[origin=c]{90}{}} & \multirow{2}[1]{*}{$T$} & \multicolumn{1}{c|}{\multirow{2}[1]{*}{Methods}} & \multicolumn{4}{c|}{Recall@$N$}    & \multicolumn{4}{c|}{Precision@$N$} & \multicolumn{4}{c|}{F1@$N$}        & \multicolumn{1}{c}{\multirow{2}[1]{*}{AUC}} \\
%		\cline{4-15}
		&       &       & $N=$1     & $N=$2     & $N=$5     & $N=$10    & $N=$1     & $N=$2     & $N=$5     & $N=$10    & $N=$1     & $N=$2     & $N=$5     & $N=$10    &  \\
		\midrule
		\multirow{6}[8]{*}{\rotatebox[origin=c]{90}{AMiner-Top}} & \multirow{2}[4]{*}{2013} & \camel & 0.1808 & 0.3035 & 0.5012 & 0.6646 & 0.3155 & 0.2734 & 0.1887 & 0.1244 & 0.2299 & 0.2877 & 0.2742 & 0.2096 & 0.8854 \\
%		&       & \proposed$_{\text{np}}$ & 0.2315&0.3708&0.5946&\textbf{0.7430}&0.3977&0.3301&0.2220&\textbf{0.1390}&0.2927&0.3493&0.3233&\textbf{0.2342}&\textbf{0.8998} \\ 		
		&       & \proposed & \textbf{0.2677} & \textbf{0.4131}& \textbf{0.6037} & \textbf{0.7220} & \textbf{0.4496} & \textbf{0.3697} & \textbf{0.2251} & \textbf{0.1360} & \textbf{0.3356} & \textbf{0.3902} & \textbf{0.3279} & \textbf{0.2289} & \textbf{0.8935} \\ 		
		\cmidrule{3-16}
		&       & Improve. & 48.06\% & 36.11\% & 20.45\% & 8.64\% & 42.50\% & 35.22\% & 19.29\% & 9.32\% & 45.98\% & 35.63\% & 19.58\% & 9.21\% & 0.91\% \\
		\cmidrule{2-16}          & \multirow{2}[4]{*}{2014} & \camel & 0.1624 & 0.2739 & 0.4831 & 0.6619 & 0.3372 & 0.2865 & 0.2094 & 0.1440 & 0.2192 & 0.2801 & 0.2922 & 0.2365 & 0.8909 \\
%		&       & \proposed$_{\text{np}}$ & 0.2146&0.3432&0.5498&\textbf{0.6967}&0.4262&0.3549&0.2381&0.1504&0.2855&0.3489&0.3322&\textbf{0.2474}&\textbf{0.8939} \\ 		
		&       & \proposed & \textbf{0.2312} & \textbf{0.3670} & \textbf{0.5679} & \textbf{0.6900}  & \textbf{0.4515} & \textbf{0.3759} & \textbf{0.2433} & \textbf{0.1507} & \textbf{0.3058} & \textbf{0.3714} & \textbf{0.3406} & \textbf{0.2473} & \textbf{0.8934} \\
		\cmidrule{3-16}          &       & Improve. & 42.36\% & 33.99\% & 17.55\% & 4.25\% & 33.90\% & 31.20\% & 16.19\% & 4.65\% & 39.51\% & 32.60\% & 16.56\% & 4.57\% & 0.28\% \\
		\bottomrule
		%	\midrule
		%    \multirow{5}[7]{*}{\rotatebox[origin=c]{90}{AMiner-All}} & \multirow{2}[4]{*}{2013} & \camel &       &       &       &       &       &       &       &       &       &       &       &       &  \\
		%          &       & \proposed &       &       &       &       &       &       &       &       &       &       &       &       &  \\
		%\cmidrule{3-16}          &       & Improve. &       &       &       &       &       &       &       &       &       &       &       &       &  \\
		%\cmidrule{2-16}          & \multirow{3}[3]{*}{2014} & \camel &       &       &       &       &       &       &       &       &       &       &       &       &  \\
		%          &       & \proposed &       &       &       &       &       &       &       &       &       &       &       &       &  \\
		%\cmidrule{3-16}          &       & Improve. &       &       &       &       &       &       &       &       &       &       &       &       &  \\
	\end{tabular}%
	\label{tab:inactive}%
	\vspace{-1ex}
\end{table*}%

\medskip
\noindent\textbf{Experimental Settings. }
We use papers published before timestamp $T$ for training, and split papers that are published after timestamp $T$ in half to make validation and test datasets. We report the test performance when the performance on validation data gives the best result, which is different from~\cite{zhang2018camel} that only has test datasets. For reliability, the reported results are averaged over 5 runs. 
Following~\cite{zhang2018camel}, we simulated 5 walks from every node, and each walk is of length 20.
We set the node embedding dimension $K=128$, pair embedding dimension $d=100$, margin $\xi=0.1$,  window size $\tau=3$, dropout ratio to 0.15, and number of negative contexts $k=1$. 
To show that~\proposed~is a general framework that is not dependent on the selection of meta-paths, we only use a single meta-path ``APA'' for~\proposed, and compare with our baseline heterogeneous network embedding methods, i.e., metapath2vec++, HNE and~\camel, that leverage multiple meta-paths, i.e., ``APA'', ``APPA'', and ``APVPA''. This is meaningful because useful meta-paths are usually task-dependent, and by relying on only a single meta-path, we demonstrate that our framework can be easily applied to various tasks. Following the setting of HNE and~\camel, we randomly sample a set of negative authors and combine it with the set of true authors to generate 100 candidate authors for each paper. For completeness, we also show evaluations on the whole authors set. For training efficiency, we pretrained the embeddings of~\proposed~with those of~\camel, but the accuracy is similar without the pretraining. The source code of~\proposed~is available on github\footnote{https://github.com/pcy1302/TapEM}.
%%\begin{table}[t]
%%	\centering
%%	\small
%%	\caption{Engineered features for supervised methods.}
%%	\vspace{-2ex}
%%	\begin{tabular}{c||c}
%%		\multicolumn{1}{c||}{\textbf{No.} } & \textbf{Feature description} \\
%%		\hline
%%		1     & paper number of the author \\
%%		2     & distinct venue number of the author \\
%%		3     & number of the paper’s references being cited by the author before \\
%%		4     & ratio of the paper’s references being cited by the author before \\
%%		5     & ratio of the author’s citations in the paper’s references \\
%%		6     & number of paper’s references in the author’s previous publications \\
%%		7     & ratio of the paper’s references in the author’s previous papers \\
%%		8     & ratio of the author’s publications in the paper’s references \\
%%		9     & number of common keyword between author and paper \\
%%		10    & ratio of the author’s keywords in common keywords \\
%%		11    & ratio of the paper’s keywords in common keywords \\
%%		12    & whether the author attend the paper’s venue before \\
%%		13    & number of times the author attend the paper’s venue before \\
%%		14    & ratio of times the author attend the paper’s venue before \\
%%		15    & number of the author’s papers in 3 years before the paper’s time \\
%%		16    & ratio of the author’s papers in 3 years before the paper’s time \\
%%	\end{tabular}%
%%	\label{tab:features}%
%%\end{table}%


\vspace{-1ex}
\subsection{Performance Analysis}
\label{sec:exp}
\textbf{RQ 1) Author identification performance}: 
Table~\ref{tab:overall} shows the author identification result of all the compared methods on both datasets in terms of various ranking metrics. We have the following observations. 
1)~\proposed~outperforms the baseline methods, especially when $N$ is small, where $N$ is the number of authors in the predicted list. This verifies that our pair embedding together with the pair validity classifier can capture the fine-grained pairwise relationship between two nodes, which pushes true authors to the top ranks. 
2)~\proposed$_{\text{npv}}$, which is a variant of~\proposed~without the pair validity classifier, still outperforms other baselines. This verifies the benefit of our pair embedding framework itself over the skip-gram based node embedding methods.
3) From the comparisons between~\proposed~and~\proposed$_{\text{npv}}$, we can verify that the pair validity classifier further improves the performance by encoding pair validity information into the pair embedding. Moreover, although not shown in the paper, it is important to note that~\proposed~converges about 10 times faster than~\proposed$_{\text{npv}}$ in average, which shows another benefit of explicitly incorporating the pair validity classifier.
4) The performance of~\proposed~is rather similar to that of~\camel~in terms of AUC, which is a metric that treats a mistake in the higher part of the ranked list as equal to one the lower part. 
Outperforming considerably in terms of position-aware metrics (e.g., Recall@N) while performing similar in terms of AUC implies that~\proposed~focuses on the accuracy of the top-ranked authors at the expense of the accuracy in the lower part of the list, which is a desideratum for a ranking algorithm.
%It is worth noting that the performance of~\proposed in terms of AUC performs on par with~\camel, which implies that the overall list is 
5) Embedding based methods generally perform better than the supervised learning--based method, indicating that the feature engineering process is error-prone. 
6) Although metapath2vec++ is a general purpose embedding method, it generally outperforms HNE, which is specifically designed for author identification task. We attribute such result to the fact that metapath2vec++ is modified to integrate the paper content, while HNE only considers the keyword of a paper. This implies that paper content plays a critical role in identifying authors. 
7) Table~\ref{tab:wholeauthor} shows the performance on the whole author candidate set of AMiner-Top dataset. We observe that the improvement of~\proposed~is more significant than the performance on the sampled author candidate set shown in Table~\ref{tab:overall}, which reaffirms the effectiveness of~\proposed. 
%\textcolor{red}{6) explain TGHNE np}

% Table generated by Excel2LaTeX from sheet 'Sheet2'
\begin{table}[t]
	\centering
	\small
	\caption{Result comparisons on the whole authors set of AMiner-Top ($T$=2013) (over all authors and inactive authors).}
	\renewcommand{\arraystretch}{0.7}
	\begin{tabular}{c|cc|c||cc|c}
		\toprule
		\multicolumn{7}{c}{Whole authors set} \\
		\midrule
		\multirow{2}[3]{*}{\begin{tabular}[x]{@{}c@{}}Recall\\@N\end{tabular}} & \multicolumn{3}{c||}{All authors} & \multicolumn{3}{c}{Inactive authors (\#papers $\leq$ 5)} \\
		\cmidrule{2-7}          & \camel & \proposed & Impr. & \camel & \proposed & Impr. \\
		\midrule
		N=10    & 0.0502 & \textbf{0.1018} & 102.79\% & 0.0304 & \textbf{0.0749} & 146.38\% \\
		N=30    & 0.1011 & \textbf{0.1955} & 93.37\% & 0.0631 & \textbf{0.1438} & 127.89\% \\
		N=50    & 0.1483 & \textbf{0.2485} & 67.57\% & 0.0943 & \textbf{0.1871} & 98.41\% \\
		N=100   & 0.2238 & \textbf{0.3520} & 57.28\% & 0.1461 & \textbf{0.2755} & 88.57\% \\
		N=200   & 0.3176 & \textbf{0.4587} & 44.43\% & 0.2213 & \textbf{0.3651} & 64.98\% \\
		\bottomrule
	\end{tabular}%
	\label{tab:wholeauthor}%
	\vspace{-2ex}
\end{table}%

\medskip
\noindent\textbf{RQ2) Performance on less active authors}: 
Recall that the skip-gram based model is not appropriate for our task because it is biased to active authors (refer to our discussions on Figure~\ref{fig:camel_walk}), and thus performs poorly on inactive authors.
However, we observe that most authors publish only few papers: in AMiner-Top and AMiner-All datasets, authors who published fewer than six papers constitute approximately 92\%, which indicates that most authors are inactive. Therefore, identifying inactive authors in author identification task is in fact crucial, but challenging owing to the limited number of historical data.
Table~\ref{tab:inactive} shows the performance comparisons on relatively inactive authors, where we consider an author inactive if he/she has fewer than 6 publications. We observe that 1)~\proposed~outperforms~\camel, and the improvement is more significant than when considering all authors: for $T$=2013, the improvement is 15.33\% on all authors but 20.45\% on inactive authors in terms of Recall@5. 
%2)~\proposed$_{\text{np}}$ performs relatively better as $N$ becomes larger, which agrees with our observation in Table~\ref{tab:overall}.
2) In Table~\ref{tab:wholeauthor} we also show that~\proposed~considerably outperforms~\camel~when the candidate authors are the set of whole authors. From the above results on less active authors, we ascertain the effectiveness of the pair validity classifier in discriminating less active true authors.

\medskip
\noindent\textbf{RQ2-1) Qualitative analysis}: 
%\noindent\textbf{RQ 3) Solution to the limitation of skip-gram based model}: 
%Recall that skip-gram based model is not appropriate for our task because it inadvertently makes a paper embedding and an author embedding similar to each other if they frequently appear together within a context window of a random walk sequence, whether or not the author is a true author of the paper (refer to our discussions on Figure~\ref{fig:camel_walk}). 
%To show the effectiveness of~\proposed~in this respect, 
To further demonstrate the effectiveness of~\proposed~in identifying inactive authors, we conduct qualitative experiment on three papers published in 2006 (AMiner-Top 2013). Our goal here  is to show that~\camel, which is based on the skip-gram model, is indeed trained to be biased to active authors, whether or not they are the true author, because an active author is more likely to appear frequently together with a target paper within the same context window of random walks.

\begin{table}[t]
	\caption{Ranking of true authors and frequently appearing false authors for a query paper. Frequently appearing false authors denote authors that frequently appear together with the query paper within the same context window of random walks, but who are not the true authors.}
	\vspace{-2ex}
	\begin{subtable}{1\linewidth}
		%		\sisetup{table-format=-1.2}
		\centering
		\small
		\renewcommand{\arraystretch}{0.7}
		\caption{Case 1: True authors contain an active author.}
		\begin{tabular}{c|c|c|c}
			\hline
			\multicolumn{4}{c}{\begin{tabular}[x]{@{}c@{}}Paper: (CIKM'06) Mining compressed commodity workflows \\from massive RFID datasets\end{tabular}}\\
			\hline
			\multirow{2}[1]{*}{} & \multirow{2}[1]{*}{\begin{tabular}[x]{@{}c@{}}Author\\(num. publications)\end{tabular}} & \multicolumn{2}{c}{Rank} \\
			\cline{3-4}
			& \multicolumn{1}{c|}{} & \multicolumn{1}{c|}{\camel} & \multicolumn{1}{c}{\proposed} \\
			\hline
			\multicolumn{1}{c|}{\multirow{3}[0]{*}{\begin{tabular}[x]{@{}c@{}}True authors\end{tabular}}} & \textbf{Jiawei Han} (141) & 1     & 8 \\
			& Xiaolei Li (12) & 198   & 1\\
			& Hector Gonzalez (9) & 296   & 81\\
			\hline
			\multicolumn{1}{c|}{\multirow{3}[0]{*}{\begin{tabular}[x]{@{}c@{}}Frequently appearing\\false authors\end{tabular}}} & Yizhou Sun (23) & 94    & 418 \\
			& Jae-Gil Lee (10) & 323   & 196 \\
			& John Paul Sondag (1) & 1043  & 3650 \\
		\end{tabular}%
		\label{tab:case1}
	\end{subtable}
	\vspace{-1ex}
	\begin{subtable}{1\linewidth}
		%		\sisetup{table-format=-1.2}
		\vspace{2ex}
		\centering
		\small
		\renewcommand{\arraystretch}{0.7}
		\caption{Case 2: Frequently appearing authors contain an active author.}
		\begin{tabular}{c|c|c|c}
			\hline
			\multicolumn{4}{c}{\begin{tabular}[x]{@{}c@{}}Paper: (KDD'06) A mixture model for contextual text mining\end{tabular}}\\
			\hline
			\multirow{2}[1]{*}{} & \multirow{2}[1]{*}{\begin{tabular}[x]{@{}c@{}}Author\\(num. publications)\end{tabular}} & \multicolumn{2}{c}{Rank} \\
			\cline{3-4}
			& \multicolumn{1}{c|}{} & \multicolumn{1}{c|}{\camel} & \multicolumn{1}{c}{\proposed} \\
			\hline
			\multicolumn{1}{c|}{\multirow{2}[1]{*}{\begin{tabular}[x]{@{}c@{}}True authors\end{tabular}}} & Cheng Xiang Zhai (51) & 4     & 3 \\
			& Qiaozhu Mei (21) & 24    & 4 \\
			\hline
			\multicolumn{1}{c|}{\multirow{2}[0]{*}{\begin{tabular}[x]{@{}c@{}}Frequently appearing\\false authors\end{tabular}}} & \textbf{Jiawei Han} (141) & 2     & 122 \\
			& Yintao Yu (6) & 601   & 372 \\
		\end{tabular}%
		\label{tab:case2}
	\end{subtable}
	\vspace{-1ex}	
	\begin{subtable}{1\linewidth}
		%		\sisetup{table-format=-1.2}
		\vspace{2ex}
		\centering
		\small
		\renewcommand{\arraystretch}{0.7}
		\caption{Case 3: Both author groups contain an active author.}
		\begin{tabular}{c|c|c|c}
			\hline
			\multicolumn{4}{c}{\begin{tabular}[x]{@{}c@{}}Paper: (KDD'06) Generating semantic annotations for\\ frequent patterns with context analysis\end{tabular}}\\
			\hline
			\multirow{2}[1]{*}{} & \multirow{2}[1]{*}{\begin{tabular}[x]{@{}c@{}}Author\\(num. publications)\end{tabular}} & \multicolumn{2}{c}{Rank} \\
			\cline{3-4}
			& \multicolumn{1}{c|}{} & \multicolumn{1}{c|}{\camel} & \multicolumn{1}{c}{\proposed} \\
			\hline
			\multicolumn{1}{c|}{\multirow{3}[0]{*}{\begin{tabular}[x]{@{}c@{}}True authors\end{tabular}}}  
			& \textbf{Jiawei Han} (141) & 1    & 14\\
			& Qiaozhu Mei (21) & 44   & 9 \\
			& Dong Xin (20) & 130   & 26 \\
			\hline
			\multicolumn{1}{c|}{\multirow{3}[0]{*}{\begin{tabular}[x]{@{}c@{}}Frequently appearing\\false authors\end{tabular}}} &\textbf{ Philip S.Yu} (122) & 7    & 41 \\
			& Xifeng Yan (36) & 15    & 19 \\
			& Charu C.Aggarwal (30) & 16   & 303\\
		\end{tabular}%
		\label{tab:case3}
%		\vspace{-1ex}
	\end{subtable}
	
	% Table generated by Excel2LaTeX from sheet 'Sheet2'
	%	\begin{subtable}{1\linewidth}
	%		%		\sisetup{table-format=-1.2}
	%		\vspace{2ex}
	%		\centering
	%		\small
	%		\renewcommand{\arraystretch}{0.8}
	%		\caption{Case 3: Both author groups contain an active author.}
	%		\begin{tabular}{c|c|c|c}
	%			\hline
	%			\multicolumn{4}{c}{\begin{tabular}[x]{@{}c@{}}Paper: (CIKM'06) C-Cubing: Efficient Computation of \\Closed Cubes by Aggregation-Based Checking\end{tabular}}\\
	%			\hline
	%			\multirow{2}[1]{*}{} & \multirow{2}[1]{*}{\begin{tabular}[x]{@{}c@{}}Author\\(num. publications)\end{tabular}} & \multicolumn{2}{c}{Rank} \\
	%			\cline{3-4}
	%			& \multicolumn{1}{c|}{} & \multicolumn{1}{c|}{\camel} & \multicolumn{1}{c}{\proposed} \\
	%			\hline
	%			\multicolumn{1}{c|}{\multirow{3}[0]{*}{\begin{tabular}[x]{@{}c@{}}True authors\end{tabular}}}  & \textbf{Jiawei Han} (141) & 12    & 32\\
	%			& Dong Xin (20) & 104   & 2 \\
	%			& Hongyan Liu (4) & 438   & 1 \\
	%			\hline
	%			\multicolumn{1}{c|}{\multirow{3}[0]{*}{\begin{tabular}[x]{@{}c@{}}Frequently appearing\\authors (not authors)\end{tabular}}} &\textbf{ Philip S.Yu} (122) & 7     & 54 \\
	%			& Xiaolei Li (12) & 67    & 41 \\
	%			& Xiaoyong Du (5) & 522   & 65\\
	%		\end{tabular}%
	%		\label{tab:case3}
	%	\end{subtable}
	\label{tab:case}
%	\vspace{-2ex}
\end{table}%

In Table~\ref{tab:case1}, we present a case where the most active author in AMiner dataset (``Jiawei Han'' in bold) is in fact one of the true authors. In this case,~\camel~successfully ranked ``Jiawei Han'' in the first place as expected, whereas the other two relatively inactive authors are ranked far below the list. On the other hand,~\proposed~showed better ranking performance regardless of the author's activeness. Table~\ref{tab:case2} shows a case where ``Jiawei Han'', who is not a true author, appeared most frequently with the query paper.
%within the same context window of meta-path guided random walks. 
In this case,~\camel~ranked ``Jiawei Han'' higher than the true authors even though he is not a true author. This behavior is expected as~\camel~is based on the skip-gram model that is biased to active authors. On the other hand,~\proposed~again demonstrates its robustness to the activeness of authors.
%thanks to the pair embedding together with the pair validity classifier. 
The last case in Table~\ref{tab:case3} shows a case where the most active author is a true author, and the second most active author in AMiner dataset, i.e., ``Philip S. Yu'', is a frequently appearing false author. In this case,~\camel~ranked ``Philip S. Yu'' higher than ``Qiaozhu Mei'' and ``Dong Xin'' due to his activeness. Moreover, we noticed that the rankings of~\camel~simply follow the order of the number of publications of the authors regardless of their authorships, which is a consequence of the skip-gram model. On the other hand,~\proposed~is robust to the activeness of authors.

To quantitatively show the reliability of the above results, we compare the ranking of true authors and frequently appearing false authors of each paper, and see whether true authors are indeed ranked higher than frequently appearing false authors. More precisely, every time a frequently appearing false author is ranked higher than a true author, we count it as a violation. For each paper, we sampled the top-N most frequently appearing false authors for comparisons, where N is the number of the actual authors. Table~\ref{tab:violation} shows the comparisons of average rank violation count. We observe that the average rank violation count of~\proposed~is about 8 times less than that of~\camel, which again demonstrates the effectiveness of our pair embedding framework.

%% Table generated by Excel2LaTeX from sheet 'Sheet2'
%\begin{table}[h]
%	\centering
%	\caption{Result for ablations of~\proposed.}
%	\begin{tabular}{r|l|cccc}
%		&    Ablations   & \multicolumn{1}{l}{Rec@1} & \multicolumn{1}{c}{Rec@2} & \multicolumn{1}{c}{Rec@5} & \multicolumn{1}{c}{Rec@10} \\
%		\midrule
%		\multicolumn{1}{c|}{\multirow{7}[0]{*}{\rotatebox[origin=c]{90}{\begin{tabular}[x]{@{}c@{}}AMiner-Top\\($T$=2013)\end{tabular}}}} & \proposed$_{\text{ndp}}$ & 0.2979 & 0.4515 & 0.6567 & 0.7766 \\
%		& \proposed$_{\text{nat}}$ &       &       &       &  \\
%		& \proposed$_{\text{nln}}$ &       &       &       &  \\
%		& \proposed$_{\text{nml}}$ &       &       &       &  \\
%		& \proposed$_{\text{npv}}$ & 0.0405      &  0.0638     &  0.1235     & 0.1926 \\
%		& \proposed$_{\text{np}}$ & 0.2780 & 0.4307 & 0.6614 & 0.7829 \\
%		& \proposed & \textbf{0.3118} & \textbf{0.4823} & \textbf{0.6807} & \textbf{0.7849} \\
%	\end{tabular}%
%	\label{tab:ablations}%
%\end{table}%

\begin{table}[t]
	\centering
	\small
	\caption{Average rank violation comparisons.}
	\renewcommand{\arraystretch}{0.85}
	\begin{tabular}{c|c|c}
		& \multicolumn{1}{c|}{\camel} &  \multicolumn{1}{l}{\proposed} \\
		\midrule
		{\begin{tabular}[x]{@{}c@{}}Average rank violation\\(the smaller the better)\end{tabular}}  & 3.8374 & 0.4519 \\
	\end{tabular}%
	\label{tab:violation}%
	\vspace{-1ex}
\end{table}%

% Table generated by Excel2LaTeX from sheet 'Sheet2'

\medskip
\noindent\textbf{RQ 3) Ablation study}: 
To measure the impact of each component of~\proposed~on the author identification accuracy, we conduct ablation studies in Table~\ref{tab:ablations}. We have the following observations:
1) Each component of~\proposed, i.e., Dropout, and attentive pooling, contributes to the performance of~\proposed.  
2) A simple linear combination of a context path instead of modeling it with a $\textsf{BiGRU}$ as in Eqn.~\ref{eqn:ctxrnn} performs worse. This implies that it is helpful to model a context path as a sequence.
3) Nevertheless, even without these components,~\proposed~still considerably outperforms the strongest baseline on AMiner-Top, which is~\camel, implying the superiority of our novel pair embedding framework.
4)~\proposed$_{\text{pv}+\text{dot}}$ is equivalent to~\proposed, but the only difference is that the final prediction is done by a dot product between paper and author embeddings as done in~\camel, instead of by the output of the pair validity classifier. We observe that~\proposed$_{\text{pv}+\text{dot}}$ still outperforms other baselines in Table~\ref{tab:overall}, which implies that the pair validity classifier is also helpful for generating more accurate paper and author embeddings.
5) We observe that~\proposed~outperforms~\proposed$_{\text{pv}+\text{dot}}$ more significantly for top-ranked authors. i.e., for small $N$ of recall@N. This implies that the pair validity classifier helps distinguish valid pairs from invalid ones, which result in pushing true authors to the top ranks. In other words, the performance for top-ranked authors suffer without the pair validity classifier.
6) The performance of $\proposed_{\text{npv}}$ that performs the worst among the ablations of~\proposed~reaffirms the benefit of the pair validity classifier.
%which is in line with the results reported in Table~\ref{tab:overall} and~\ref{tab:inactive}.

\begin{figure}[htbp]
%	\vspace{-1ex}
	\centering
	\includegraphics[width=0.9\linewidth]{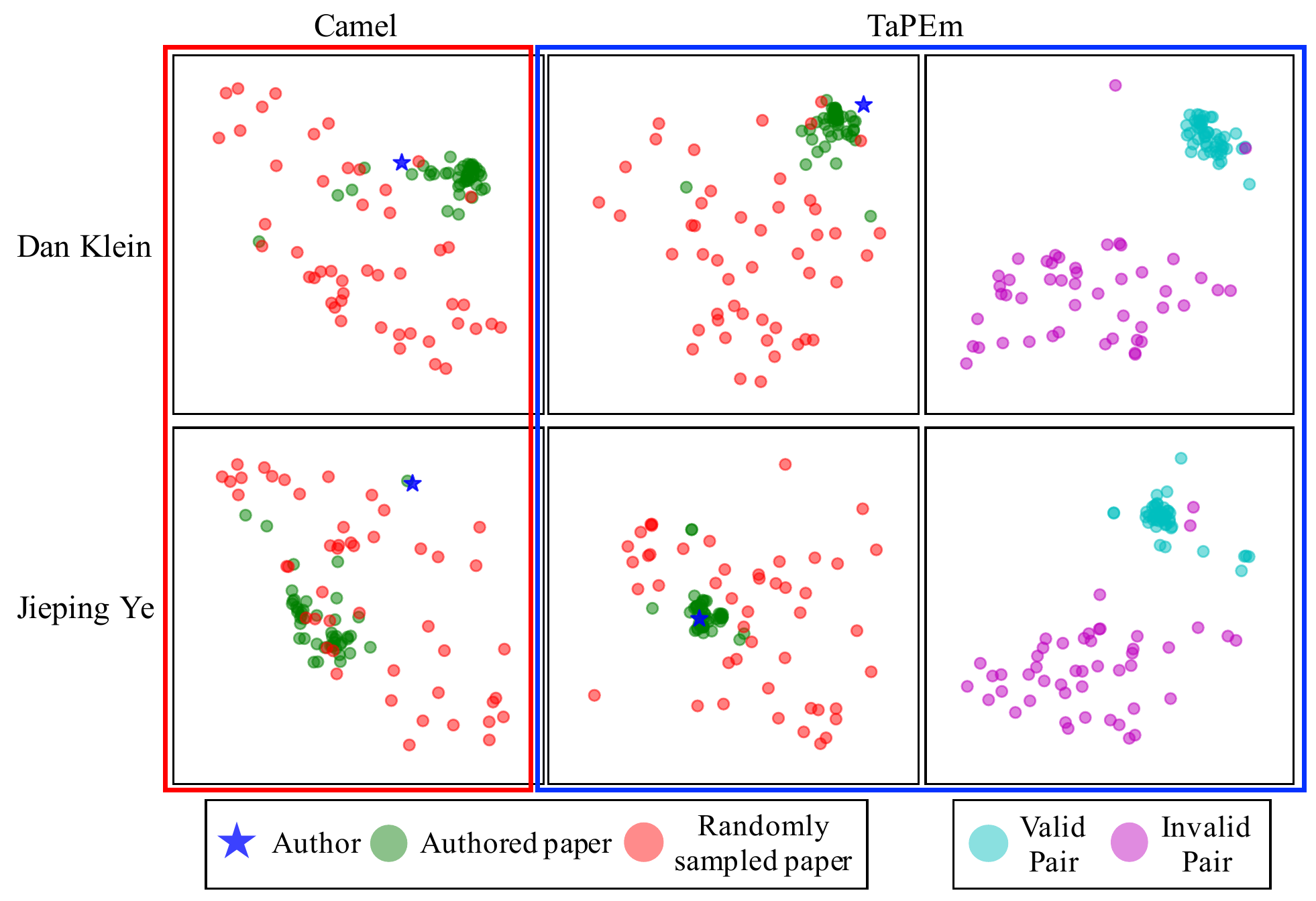}
%	\vspace{-2ex}
	\caption{t-SNE visualization of author, paper and pair embeddings for two authors: Dan Klein and Jieping Ye.}
	\label{fig:visualization}
	\vspace{-2ex}	
\end{figure}

\begin{table}[t]
	%	\vspace{-2ex}	
	\centering
	\small
	\caption{Result for ablations of~\proposed.}
	\vspace{-2ex}
	\renewcommand{\arraystretch}{.7}
	\begin{tabular}{r|l|cccc}
		&    Ablations   & \multicolumn{1}{l}{Rec@1} & \multicolumn{1}{c}{Rec@2} & \multicolumn{1}{c}{Rec@5} & \multicolumn{1}{c}{Rec@10} \\
		\midrule
		\multicolumn{1}{c|}{\multirow{6}[0]{*}{{\begin{tabular}[x]{@{}c@{}}AMiner-Top\\($T$=2013)\end{tabular}}}} & No $\textsf{Dropout}$ & 0.2979 & 0.4515 & 0.6567 & 0.7766 \\
		%		\multicolumn{1}{c|}{\multirow{4}[0]{*}{\rotatebox[origin=c]{90}{\begin{tabular}[x]{@{}c@{}}AMiner-Top\\($T$=2013)\end{tabular}}}} & No $\textsf{Dropout}$ & 0.2979 & 0.4515 & 0.6567 & 0.7766 \\
		& No attention &   0.3024    &  0.4627     &   0.6679    &  0.7800\\
		& No Eqn.~\ref{eqn:ctxrnn} &   0.2824    &  0.4499     &   0.6623    &  0.7792\\		
		\cmidrule{2-6}
		%		& No LayerNorm &       &       &       &  \\
		%		& No $\mathcal{L}_{M e t r i c}$ &       &       &       &  \\
		%		& No $\mathcal{L}_{\text{pv}}$ & 0.0405      &  0.0638     &  0.1235     & 0.1926 \\
		& \proposed$_{\text{npv}}$ & 0.2755 & 0.4266 & 0.6405 & {0.7677} \\
		& \proposed$_{\text{pv}+\text{dot}}$ & 0.2858 & 0.4392 & 0.6614 & \textbf{0.7906} \\
		& \proposed & \textbf{0.3118} & \textbf{0.4823} & \textbf{0.6807} & {0.7849} \\
	\end{tabular}%
	\label{tab:ablations}%
	\vspace{-1ex}
\end{table}%

\medskip
\noindent\textbf{RQ 4) Visualization of embeddings: } 
To provide a more intuitive understanding of pair embeddings, we visualize paper, author embeddings and paper--author pair embeddings of two authors from AMiner-Top dataset by using t-SNE~\cite{maaten2008visualizing}. More precisely, for~\camel, we plot an author along with the papers written by the author, and we also plot randomly sampled papers that are not written by the author as many as the number of authored papers. We also plot both paper/author embeddings, and pair embeddings of~\proposed. Note that the same set of authored papers and randomly sampled papers used for~\camel~are used for constructing the pair embeddings.

Compared with~\camel, we observe in Figure~\ref{fig:visualization} that the embeddings of authored papers of~\proposed~are more tightly grouped together than those of~\camel, and the author embedding of~\proposed~is placed relatively closer to the cluster of the authored papers than those of~\camel. 
This implies that~\proposed~generates more accurate representations of paper and author than~\camel~by making two nodes that constitute a pair close to each other not only if they are related to a similar research topic, but also the pair itself is valid at the same time;
%the pair validity classifier of~\proposed \\makes two nodes that constitute a pair close to each other not only if they are related to a similar research topic, but also the pair itself is valid at the same time, which in turn generates more accurate representations of paper and author;
%generates more accurate representations of paper and author, thanks to the pair embedding scheme; 
this is corroborated by the result of~\proposed$_{\text{pv}+\text{dot}}$ in Table~\ref{tab:ablations} that outperforms~\camel~in Table~\ref{tab:overall}. Moreover, we observe from the rightmost figure that when an author is coupled with the papers (both authored papers and randomly sampled papers) to form paper--author pair embeddings, it becomes easier to distinguish whether a pair is valid or not, which is the benefit obtained from the pair validity classifier. This is a useful for task-guided heterogeneous network embedding whose ultimate goal is to model the likelihood of pairwise relationship between two nodes.

\subsubsection{Discussion}
In the experiments, we focused on the problem of author identification as an application of our proposed framework.  Under the space limitation, our intention is to delve deep into showing the effectiveness of~\proposed~in various aspects, instead of covering many tasks but with limited experiments.
However, we postulate that the final objective function $\mathcal{L}$ in Eqn.~\ref{eqn:final} can be leveraged to solve various real-world tasks whose the ultimate goal is to model the  pairwise relationship between node $v$ and $u$; user-item relationship in recommendation, paper-paper relationship in citation recommendation, and author-author relationship in collaborator recommendation. 

\section{Conclusion}
In this paper, we proposed a novel task-guided pair embedding framework in heterogeneous network embedding that is useful for tasks whose goal is to predict the likelihood of pairwise relationship between two nodes. Instead of learning general purpose node embeddings, we directly focus on the pairwise relationship between two nodes that we are interested in, and learn the pair embedding considering its associated context path between the pair of nodes. 
Our pair validity classifier is effective in identifying less active true authors, and pushing true authors to the top ranks, which is desideratum for a ranking algorithm. As future work, we plan to investigate on the applicability of~\proposed~on different tasks.
%Moreover, our pair validity classifier has shown its effectiveness in pushing true authors to the top ranks, which is desideratum for a ranking algorithm. As our future work, we plan to investigate on the applicability of~\proposed~on different tasks.
%Although we focused on the author identification task in this paper, our framework can be readily adopted to various tasks aiming at predicting the likelihood between two nodes, which we leave as the future work.

\noindent\textbf{Acknowledgment}:
2016R1E1A1A01942642, and SW Starlab (IITP-2018-0-00584).
%\newpage
\bibliographystyle{ACM-Reference-Format}
\bibliography{sample-sigconf}

\end{document}